\documentclass{article}

\usepackage{microtype}
\usepackage{graphicx}
\usepackage{subcaption}
\usepackage{caption}
\usepackage[normalem]{ulem}
\usepackage{wrapfig}
\usepackage[htt]{hyphenat}
\usepackage{mdframed}
\usepackage{tcolorbox}
\makeatletter
\newcommand{\disablepackage}[2]{%
  \disable@package@load{#1}{#2}%
}
\newcommand{\reenablepackage}[1]{%
  \reenable@package@load{#1}%
}
\makeatother

\usepackage{booktabs}
\usepackage{array}
\usepackage{makecell}

\usepackage{svg}
\graphicspath{ {./images/} }

\usepackage[disable]{todonotes}
\usepackage{enumitem}

\usepackage{hyperref}

\usepackage[preprint]{icml2026}
\renewcommand{\algorithmiccomment}[1]{\hfill$\triangleright$ #1}

\usepackage{amsmath}
\usepackage{amssymb}
\usepackage{mathtools}
\usepackage{amsthm}
\usepackage{dsfont}
\usepackage{mathtools}
\providecommand{\abs}[1]{\lvert#1\rvert}
\providecommand{\norm}[1]{\lVert#1\rVert}
\DeclareMathOperator*{\argmax}{arg\,max\,}
\DeclareMathOperator*{\argmin}{arg\,min\,}

\usepackage{appendix}

\theoremstyle{plain}
\newtheorem{theorem}{Theorem}[section]

\theoremstyle{definition}

\theoremstyle{remark}

\icmltitlerunning{Information Templates: A New Paradigm for Intelligent Active Feature Acquisition}

\begin{document}

\twocolumn[
  \icmltitle{Information Templates: A New Paradigm for Intelligent \\ Active Feature Acquisition}



  \icmlsetsymbol{equal}{*}

  \begin{icmlauthorlist}
    \icmlauthor{Hung-Tien Huang,}{equal,uncch-cs}
    \icmlauthor{Dzung Dinh}{equal,uncch-cs}
    \icmlauthor{Junier B. Oliva}{uncch-cs}
  \end{icmlauthorlist}

  \icmlaffiliation{uncch-cs}{Department of Computer Science, University of North Carolina at Chapel Hill, Chapel Hill, U.S.A}

  \icmlcorrespondingauthor{Hung-Tien Huang}{hungtien@unc.edu}

  \icmlkeywords{Machine Learning, ICML}

  \vskip 0.3in
]



\printAffiliationsAndNotice{\icmlEqualContribution}

\begin{abstract}
  Active feature acquisition (AFA) is an instance-adaptive paradigm in which, at inference time, a policy sequentially chooses which features to acquire (at a cost) before predicting. Existing approaches either train reinforcement learning policies, which deal with a difficult MDP, or greedy policies that cannot account for the joint informativeness of features or require knowledge about the underlying data distribution. To overcome this, we propose Template-based AFA (TAFA), a non-greedy framework that learns a small library of feature templates---sets of features that are jointly informative---and uses this library of templates to guide the next feature acquisitions. Through identifying feature templates, the proposed framework not only significantly reduces the action space considered by the policy but also alleviates the need to estimate the underlying data distribution. Extensive experiments on synthetic and real-world datasets show that TAFA outperforms the existing state-of-the-art baselines while achieving lower overall acquisition cost and computation.\looseness-1
\end{abstract}

\section{Introduction}\label{sec:intro}
Many machine learning (ML) methods implicitly assume fully observed feature vectors at inference time; in practice, however, acquiring features often incurs non-trivial costs in time, money, or effort. \emph{Active feature acquisition} \citep[AFA;][]{saar-2009} addresses this mismatch by treating inference as a sequential process in which an agent adaptively selects features while balancing predictive accuracy and acquisition cost. AFA policies personalize the selected set of features from instance to instance due to their sequential nature, where future acquisitions depend on the observed values of previous acquisitions. For example, in health care applications, this means that certain patients may have their demographic, lab, and X-ray related features acquired for inference, while other patients may have demographic and EEG-based features acquired. Beyond healthcare, applications abound in robotics (making decisions with judicious use of sensors), education (assessing student knowledge with judicious question choices), troubleshooting (assessing the issues with judicious choices of diagnostic tests), etc. 

Given the combinatorial nature of the dynamic AFA acquisition process, the potential acquisition space is exponential in instances' dimensionality. This large decision space, in turn, makes learning, computing, and \emph{understanding} policies difficult. In this work, we turn our attention to a previously overlooked question: \emph{what are the canonical feature subsets driving the cost/benefit performance in AFA tasks?} We answer this question and introduce template-based AFA (TAFA), a novel approach that simplifies AFA according to a learned dictionary of feature subsets (\emph{templates}) in order to improve both the efficacy as well as inference time of AFA agents, all whilst avoiding the need for RL-based training. In addition, we leverage our template concept to learn interpretable policies that succinctly and exactly characterize the subsets of features an AFA agent may acquire, as well as the types of instances that map to each subset—capturing both \emph{what information matters} and \emph{for whom}.

\textbf{Contributions}\quad \uline{First}, we introduce a template-based framework for efficient non-greedy acquisition, achieving strong accuracy-cost tradeoffs with substantially lower inference latency. \uline{Second}, we develop a mutation-greedy search to construct high-quality template libraries, and an end-to-end refinement stage that tunes templates via Gumbel-Softmax relaxation \citep{jang2016categorical}. \uline{Third}, we provide theoretical analysis showing the template search objective is submodular (yielding approximation guarantees), and our cost/benefit criteria form a lower bound on the optimal AFA MDP value. \uline{Fourth}, we present a student/teacher distillation framework that yields an interpretable decision-tree policy with a human-readable list of possible acquisition subsets and corresponding rules. \uline{Fifth}, we conduct extensive experiments demonstrating that TAFA outperforms baselines in both predictive performance and computational efficiency.

\section{AFA Background and Related Works}\label{sec:problem}
\textbf{Notation and Data}\quad Let $x = \langle x_d \rangle_{d=1}^{D} \in \mathbb{R}^D$ be a $D$-dimensional input feature vector and $y$ be its corresponding output for a supervised learning task. We use $\mathbf{o} \subseteq \left[D\right] \equiv \{1,\cdots, D\}$ and $\mathbf{o} \in \mathfrak{P}\left(\left[D\right]\right)$ to denote a subset of feature indices, where $\mathfrak{P}(\cdot)$ is the power set. Let $x_{\mathbf{o}} \in \mathbb{R}^{\abs{\mathbf{o}}}$ be the corresponding subset feature values indexed by $\mathbf{o}$. We assume standard access to a supervised learning dataset \scalebox{.9}{$\mathcal{D}_{\text{trn}} = \left\{ (x^{(n)}, y^{(n)}) \right\}_{n=1}^{N} \overset{\mathrm{iid}}{\sim} \mathcal{P}$}, where $\mathcal{P}$ is the data distribution, and a predictor $\widehat{y}(x_{\mathbf{o}})$ that can make an inference about the response variable using an arbitrary subset of features from the corresponding instance.

\begin{figure}[h]
    \centering
    \includegraphics[width=\linewidth]{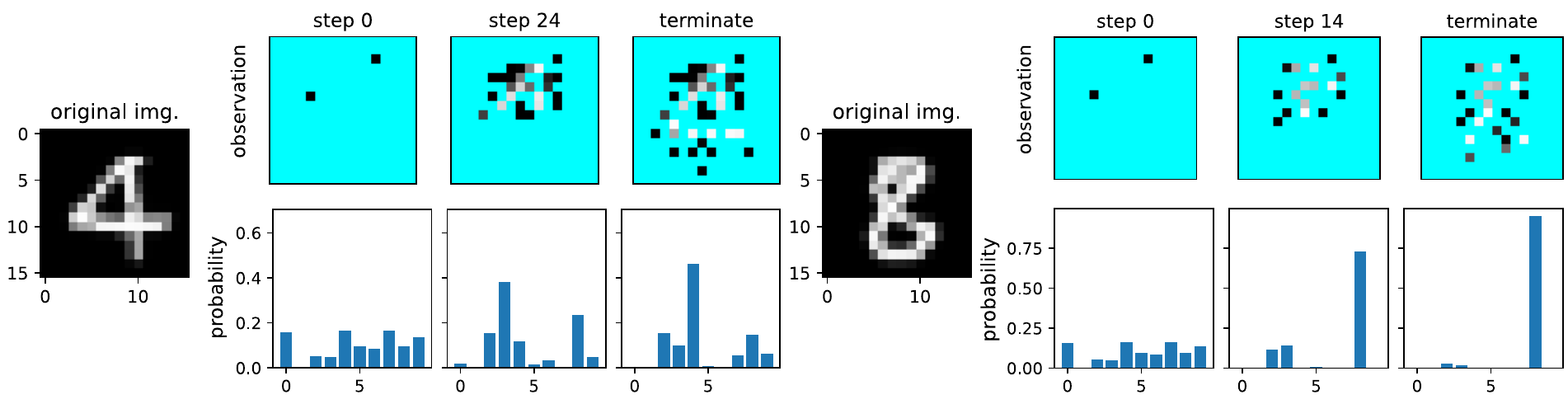}
    \caption{Example of rolling out an AFA policy. The AFA agent continues gathering information until it has enough information to make a decision about the instance at hand.}
    \label{fig:mnist-rollout-sample}
\end{figure}

\textbf{Sequential Acquisition and Policies}\quad AFA~\citep{rahb-2025} formulates inference as a sequential process in which an agent acquires unobserved feature values at a cost and decides when to stop acquisition and make a prediction. That is, for a current set of observed features $\mathbf{o}$, an AFA policy $\pi(x_{\mathbf{o}}) \in \left(\left[ D \right] \backslash \mathbf{o}\right) \cup \left\{ \varnothing \right\}$, either: 1) acquires a new feature $\mathbf{o} \leftarrow \mathbf{o} \cup \{\mathbf{a}\}$ when choosing action $a = \pi(x_{\mathbf{o}}) \in \left[ D \right]$ (i.e., continuing the policy at $\pi(x_{\mathbf{o}\cup\{a\}})$); or 2) uses the predictor to infer $\widehat{y}(x_{\mathbf{o}})$ when terminating action is chosen $a = \pi(x_{\mathbf{o}}) = \varnothing$ (see \autoref{fig:mnist-rollout-sample}). \looseness-1

\textbf{Conditional Mutual Information}\quad Several approaches learn AFA policies via conditional mutual information (CMI) estimation, greedily selecting the next feature expected to yield the greatest information gain, often by training generative models to approximate the data distribution $\mathcal{P}$ \cite{ma-2019}. As \citet{gadg-2024} discussed, while effective, generative modeling can introduce practical challenges in terms of computational cost and optimization complexity. In a complementary direction, \citet{gadg-2024} proposed discriminative mutual information estimation (DIME), which recovers CMI at optimality through a carefully designed objective using a regression model, without requiring a generative model.

\textbf{Reinforcement Learning Approaches}\quad The sequential active acquisition problem can be posed as a Markov Decision Process \citep[MDP;][]{dula-2011,shim-2018,li-2021} where states are observed features and actions correspond to either acquiring an unobserved feature or terminating the acquisition process; the rewards can reflect negative acquisition costs and predictive performance measures (when terminating). After deciding to terminate acquisition and use a feature subset $\mathbf{b} \subseteq \left[D\right]$ to infer about the ground truth label $y$, the cumulative reward for an AFA episode on instance $x$ is then $G = -e(x_{\textbf{b}}, y)$; we coin $e$ as the \emph{negative cumulative reward objective}:
\begin{equation}
    e(x_{\textbf{b}}, y) = l( \widehat{y} ( x_{\textbf{b}} ), y ) + \lambda \sum_{b \in \textbf{b}} c(b),  
    \label{eq:subset_feature_loss_fn}
\end{equation}

where $l$ is a supervised-learning (e.g., cross-entropy) loss, $c: \left[D\right] \mapsto \mathbb{R}_{+}$ is a relative cost associated with each feature, and $\lambda$ is an \emph{application-specific} trade-off parameter controlling the trade-off between cost and predictive performance. The negative cumulative reward objective (\autoref{eq:subset_feature_loss_fn}) shall play an integral role in the design of our AFA templates below.

\textbf{Direct (Non-RL) Acquisition Objectives}\quad Training an RL agent can overcome limitations of CMI-based methods, which may fail to capture features that are only informative jointly with respect to the target \citep{vala-2024}. However, learning RL agents for the AFA MDP remains challenging due to poor sample efficiency, training instability, catastrophic forgetting \citep{ding-2020}, large action spaces, and credit assignment difficulties \citep{li-2021}.
To avoid RL drawbacks, SEFA \citep{norcliffe2025stochastic} proposes learning feature-wise stochastic latent representations and selecting features by maximizing an expected gradient-based objective in latent space. While this yields a non-RL acquisition strategy, it requires training an additional latent model and is not directly tied to the AFA MDP reward structure.
In a complementary approach, \citet{vala-2024} introduced the acquisition-conditioned oracle (ACO), which avoids both RL and generative modeling by estimating likely values of unobserved features at inference time and selecting informative, cost-effective subsets. Despite its theoretical appeal, ACO is computationally prohibitive, as it evaluates objectives over exponentially many future acquisition sequences; even its approximate variant (AACO), which subsamples, remains slow at inference.

\emph{Our TAFA approach} yields a policy with a decision space that is constrained to a learned collection of templates and avoids planning that infers over all possible feature combinations, improving efficiency while staying aligned with AFA MDP rewards
(\autoref{thm:lower}).

\textbf{Interpretable AFA}\quad Despite their effectiveness, most AFA policies provide limited justification for why a particular instance is routed to a specific acquisition set. One recent approach \citep{guney2025active} applies post-hoc explanations (e.g., SHAP \citep{lundberg2017unified}) to obtain instance-wise feature importance rankings and then learns a sequential policy that selects the next unacquired feature with the highest attribution given the current partial information. While intuitive, this strategy trains the policy to follow the predictor's attributions rather than yielding transparent, rule-based justifications for acquisition decisions.
In parallel, works on interpretable RL have built structured policies such as decision-tree and differentiable decision-tree models \citep{silva2020ddt, marton2024sympol}. Although these approaches can be adapted to AFA by treating feature indices as actions, their single-feature action formulation makes capturing set-level interactions difficult, often requiring large trees to approach black-box performance.\looseness-1

\emph{Our TAFA approach} compresses the decision space and enables distillation into compact step-wise trees with explicit \textsc{Acquire If} rules, yielding an \emph{interpretable} policy.

\textbf{Submodularity and Feature Selection}\quad Optimal feature selection often exhibits submodular structure with diminishing marginal utility as subset size grows, enabling greedy algorithms with provable guarantees \citep{nemh-1978}. Early work leveraged these properties for \emph{dataset-level}, \uline{static} feature selection that identifies a \emph{single subset} intended to perform well on average across instances \citep{Kusner2014,Miller2002}.

\emph{Our TAFA approach} departs from the static paradigm (and is the first to exploit submodularity in AFA, to our knowledge) by learning \emph{multiple} feature subsets (templates) that enable instance-adaptive inference-time acquisition, allowing different instances to be routed to different informative subsets. The TAFA objective is submodular (\autoref{thm:submod}), preserving the guarantees of greedy optimization while capturing diminishing returns as templates are introduced.\looseness-1

\section{Methodology} \label{sec:method}
\begin{figure}[b]
    \centering
    \includegraphics[width=.85\linewidth]{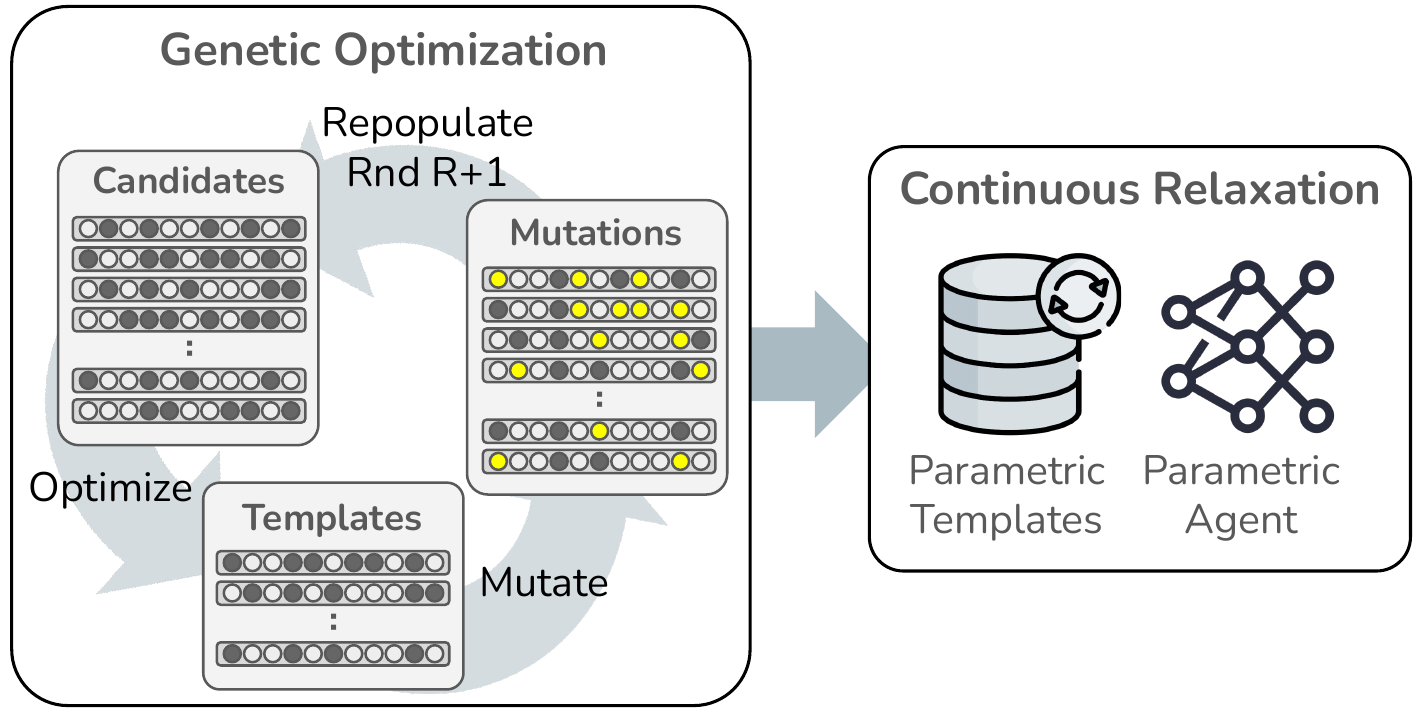}
    \caption{Our proposed template optimization procedure.}
    \label{fig:itr-muatate-comic}
\end{figure}
Our proposed template-based approach searches for an idealized collection of feature subsets, the \emph{templates}, which enable a near-optimal cost/benefit objective (\autoref{eq:subset_feature_loss_fn}) under oracle selection on training instances (\autoref{sec:template_search}). This leads to a submodular set optimization problem, for which we derive a genetic algorithm (\autoref{sec:mutate_search}; see \autoref{fig:itr-muatate-comic}). Moreover, we develop a discrete Gumbel relaxation technique to further refine templates (\autoref{sec:gumbel}). Finally, utilizing the templates, we propose an intrinsically interpretable policy that explicitly shows the policy's decision process (\autoref{sec:trees}). Our template-based policies enable faster inference with strong teachers, which in turn enables the efficient training of an interpretable student.

\subsection{Template Search as Set Optimization}
\label{sec:template_search}
We propose to find a collection $\mathcal{B}^{\star} = \{\textbf{b}^{(i)}\}_{i=1}^{B}$ of $B$ feature templates where each template $\textbf{b}^{(i)} \subseteq \left[ D \right]$ is an informative combination of features.
We formulate the problem of finding an optimal collection of templates as a set optimization problem\footnote{A function $g:\mathfrak{P}(\mathcal{V}) \rightarrow \mathbb{R}$ is said to be a set function if it assigns each subset $\mathcal{S} \subseteq \mathcal{V}$ of a finite set $\mathcal{V}$ to a value $g(\mathcal{S})$.} \todo{footnote: do you mean $g(\mathcal{S})$ }. As discussed earlier, an effective AFA policy must balance acquiring informative features against the associated costs; therefore, we define the objective function as:\looseness-1
\begin{equation}
    \scalebox{0.93}{$\displaystyle g(\mathcal{B}) = \mathbb{E}_{x,y \sim \mathcal{P}} \left[  
                \min_{\textbf{b} \in \mathcal{B}}  
                l( \widehat{y} ( x_{\textbf{b}} ), y ) + \lambda \sum_{b \in \textbf{b}} c(b)  
                \right]$}  
    \label{eq:tafa_set_obj_fn}
\end{equation}
i.e., $g(\mathcal{B}) = \mathbb{E}_{x,y \sim \mathcal{P}} \left[ \min_{\textbf{b} \in \mathcal{B}} e(x_\textbf{b}, y) \right]$, the expected oracle cost/benefit objective (\autoref{eq:subset_feature_loss_fn}) for data instances with templates in $\mathcal{B}$. Thus, our template optimization problem is:
\begin{equation}
    \mathcal{B}^{\star} = \argmin_{\mathcal{B} \in \mathfrak{S}_B }  g(\mathcal{B}) ,\quad 
    \label{eq:tafa_opt_prob}
\end{equation}
where $\mathfrak{S}_B$ represents the search space of all possible template collections, $|\mathfrak{S}_B| = {\mathcal{O}}\left({{2^D \choose B}}\right)$.
An important property of our objective function is that it exhibits submodularity under certain conditions.
\begin{theorem}
    \label{thm:submod}
    (Informal) The template collection objective function $g(\mathcal{B})$ defined in \autoref{eq:tafa_set_obj_fn} is submodular in $\mathcal{B}$.
    \begin{proof}
        See \citet{krau-2008}, the ``facility location'' problem discussed in \citet{krau-2014}, and \autoref{sec:submodularity} in the supplementary for details of the proof.
    \end{proof}
\end{theorem}
This submodularity property ensures diminishing returns: adding a new template to a larger collection provides less benefit than adding it to a smaller collection. Under additional monotonicity conditions, this guarantees that solving \autoref{eq:tafa_opt_prob} with greedy optimization algorithms achieves good approximation bounds (see \autoref{sec:greedy_optimality}~for details). However, even a greedy algorithm shall be $\mathcal{O}(B 2^D)$ and thus intractable. Hence, below we further develop evolution-based and relaxation strategies.

\subsection{Template-Based Policy}
\label{sec:template_based_policy}
Here, we discuss how one may construct AFA policies through an optimized template collection $\mathcal{B}$.
Intuitively, a straightforward policy operates in two stages: template selection and feature acquisition. Given current observations $x_{\textbf{o}}$, we first select the most promising template (\autoref{eq:subset_feature_loss_fn}):
\begin{equation}
    \textstyle
    \textbf{b}^{\star} = \argmin_{\textbf{b} \in \mathcal{B}} \left\{ \widehat{l_{\textbf{b}}}( x_{\textbf{o}} ) + \lambda \sum_{j \in \textbf{b} \backslash \textbf{o} } c(j) \right\}
    \label{eq:template_selection}
\end{equation}
where $\widehat{l_{\textbf{b}}}( x_{\textbf{o}} )$ estimates the prediction loss when using template $\textbf{b}$ given the current observations, $\mathbb{E}_{y, x_\textbf{b} \mid x_{\textbf{o}}}[l\left(\widehat{y}\left(x_{\textbf{o} \cup \textbf{b}}\right), y\right)]$\footnote{$\widehat{l_{\textbf{b}}}( x_{\textbf{o}} )$ may be estimated in a \emph{supervised fashion}, e.g., utilizing the $k$-nearest training instances based on observed features $\norm{x_{\textbf{o}} - x^{\prime}_{\textbf{o}}}_2$ and averaging the cached losses over the neighbors.}, and $\lambda \sum_{j \in \textbf{b} \backslash \textbf{o} } c(j)$ represents the remaining acquisition cost for unobserved features in template $\textbf{b}$. Once a template is selected, we acquire the cheapest unobserved feature from that template:
\begin{equation}
    \pi(x_{\textbf{o}}) = \argmin_{a \in \textbf{b}^{\star} \backslash \textbf{o}} c(a)
    \label{eq:feature_acquisition}
\end{equation}
We may show that our criteria \autoref{eq:feature_acquisition} is indicative of the optimal AFA MDP value of states.
\begin{theorem}
    \label{thm:lower}
    (Informal) The AFA MDP value function is lower bounded by the TAFA criterion:
    \begin{align*}
        V(x_{\textbf{o}})
         & \geq \max_{\textbf{b} \in \mathcal{B}}{-\mathbb{E}_{y, x_{\textbf{b}} \vert x_{\textbf{o}}}\left[ l\left(\widehat{y}\left(x_{\textbf{o} \cup \textbf{b}}\right), y\right) \right] - \lambda \sum_{u \in \textbf{b}} c(u)}
    \end{align*}
    \begin{proof}
        See \autoref{sec:tafa_value_bound}~for details.
    \end{proof}
\end{theorem}
This template-based policy can be viewed as a structured approximation to more general AFA approaches. While methods like ACO \citep{vala-2024} consider all possible feature acquisition sequences, our approach restricts choices to a learned set of templates $\mathcal{B}$, trading optimality for computational efficiency and interpretability.\looseness-1

\subsection{Mutation-Guided Greedy Template Search} \label{sec:mutate_search}
As aforementioned, the large $\mathcal{O}(2^D)$ space of potential templates makes even greedy routines intractable, notwithstanding submodularity. Thus, here we develop a genetic search procedure that conducts a greedy search on a smaller, mutated candidate set of templates, as expounded below.

\begin{algorithm}
    \caption{\textsc{IterativeMutateSearch}}  \label{alg:mutate_greedy_search}
    \small
    \begin{algorithmic}[1]
        \REQUIRE training dataset $\mathcal{D}_{\text{trn}}$, initial feature $o_{\text{init}}$, predictor $\widehat{y}$, task objective function $l$, cost function $c$, template set size $B$, candidate set size $S$, mutative rounds $R$.
        \ENSURE $\widehat{B}$ an estimate for optimal collection of templates.
        \FOR{$r \gets 0 \cdots R$}
        \IF{r\,=\,0}
        \STATE $\mathcal{C}^{(r)} \gets \textsc{MakeRandomCandidates}(D, S)$
        \ELSE
        \STATE $\mathcal{C}^{(r)} \gets \text{\textsc{Mutate}}(\widehat{\mathcal{B}}^{(r-1)}, S)$
        \ENDIF
        \STATE Evaluate $e(x_{\widetilde{\textbf{b}}}, y) \quad\forall (x, y), \widetilde{\textbf{b}} \in \mathcal{D}_{\text{trn}} \times \mathcal{C}^{(r)}$ \COMMENT{\autoref{eq:subset_feature_loss_fn}}
        \STATE Initialize \scalebox{0.8}{$\displaystyle\widehat{\mathcal{B}}_{0}^{(r)} \gets \left\{ \argmin_{\textbf{b} \in \mathcal{C}^{(r)}} \frac{1}{\abs{\mathcal{D}_{\text{trn}}}} \sum_{x, y \in \mathcal{D}_{\text{trn}}} e( x_{\textbf{b}}, y ) \right\}$}  
        \FOR{$t \in 1 \cdots B$}
        \STATE \scalebox{0.9}{$\displaystyle\textbf{b}^{\star}_t \gets \argmin_{\widetilde{\textbf{b}}}{ 
                    \sum_{x, y} {  
                        \min\left({
                            e\left( x_{\widetilde{\textbf{b}}}, y \right),
                            \min_{\textbf{b}} e\left( x_{\textbf{b}}, y \right)
                        }
                        \right)
                    }
                }$} \COMMENT{\autoref{eq:greedy_criterion}}
        \STATE $\widehat{\mathcal{B}}_{t}^{(r)} \gets \widehat{\mathcal{B}}_{t-1}^{(r)} \cup \left\{ \textbf{b}^{\star}_{t}\right\}$
        \ENDFOR
        \STATE $\widehat{\mathcal{B}}^{(r)} \gets \widehat{\mathcal{B}}_{T}^{(r)}$
        \ENDFOR
        \STATE \textbf{return} $\widehat{\mathcal{B}}^{(R)}$
    \end{algorithmic}
    \normalsize
\end{algorithm}
Given a candidate set, $\mathcal{C}$, to select templates from, a training dataset, $\mathcal{D}_{\text{trn}}$, and previously selected templates, $\widehat{\mathcal{B}}_{t-1}$, it is straightforward to show that the next template selection problem optimizes:
\begin{equation}
    \scalebox{0.9}{$\displaystyle\textbf{b}^{\star}_{t} = \argmin_{\widetilde{\textbf{b}} \in \mathcal{C} \backslash \widehat{\mathcal{B}}_{t-1}}{ 
                \frac{1}{\abs{\mathcal{D}_{\text{trn}}}} \sum_{\mathcal{D}_{\text{trn}}} {  
                    \min\left({e\left( x_{\widetilde{\textbf{b}}}, y \right),
                        \min_{\textbf{b} \in \widehat{\mathcal{B}}_{t-1}}{\!\!
                            e\left( x_{\textbf{b}}, y \right)
                        }
                    }
                    \right)
                }
            }$}
    \label{eq:greedy_criterion}
\end{equation}
Intuitively, \autoref{eq:greedy_criterion}  asks: \textit{if we add candidate $\widetilde{\textbf{b}}$ to $\mathcal{B}_{t-1}$, how much does the best-achievable performance improve on average.} Note that selection begins with the special case:  $\mathcal{C}^{(r)} = \{ \textbf{b}^{(s)} \mid \textbf{b}^{(s)} \sim \{ \textbf{b}' \mid o_{\text{init}} \in \textbf{b}' \subseteq [D] \} \}_{s=1}^{S}$\footnote{We ensure that candidate templates carry the pre-optimized initial feature $o_{\text{init}}$ (e.g., optimized with an initial search as in \citet{vala-2024}, cross-validated, or specified w/ prior knowledge).}. As an unrestricted candidate set, $\mathcal{C}$, is exponential in size, we propose a genetic, mutation-guided optimization.

Our method draws inspiration from genetic algorithms \citep{holl-1973,forr-1996,siva-2008}, where quality solutions are used to generate new candidates via mutations. The key insight is that good templates from one round likely contain informative feature combinations that, when slightly modified, can lead to better templates.

The algorithm proceeds in rounds. At each round $r$, we first run a greedy search over the current candidate template set $\mathcal{C}^{(r)}$, yielding a selected template collection $\widehat{\mathcal{B}}^{(r)}$. The candidate set for the next round is then constructed by 1) mutating each template in $\widehat{\mathcal{B}}^{(r)}$, 2) carrying forward the previously optimized templates $\widehat{\mathcal{B}}^{(r)}$, and 3) injecting an additional set of randomly sampled templates to encourage exploration. Mutations randomly drop features from existing templates with probability 0.5. This design balances exploitation—by refining high-utility templates—with exploration through stochastic mutations and fresh random candidates. The initial candidate set $\mathcal{C}^{(0)}$ is populated entirely with random templates. See \autoref{alg:mutate_greedy_search} for the pseudocode and \autoref{fig:itr-muatate-comic} for a visual.

\subsection{Template Search Through Continuous Relaxation} \label{sec:gumbel}
To further refine the template sets, we provide an additional \emph{non-greedy} tuning routine through continuous relaxation~\citep{jang2016categorical} to optimize templates w.r.t.~\emph{a learned actor} (rather than an oracle selection, \autoref{eq:tafa_set_obj_fn}), yielding an updated set of templates for deployment.

\textbf{Parameterized Templates}\quad To make templates differentiable, we compute the relaxed version of our template bank $\mathcal{B} = \{\textbf{b}^{(i)}\}_{i=1}^{B} $ as follows. Let $\widetilde{\mathcal{M}} = \langle \widetilde{m}^{(i)}\rangle_{i=1}^{B} \in \mathbb{R}^{B \times D}$ denote the relaxed feature template bank, where each of the $\widetilde{m}^{(i)}$'s are defined as $\mathrm{Sigmoid}\left(\mu_i\right)$ and $\mu \in \mathbb{R}^D$ is the template parameter. We exclude $m^{(i)}_{o_{\text{init}}}$ from back propagation to avoid optimization from accidentally turning off the initial feature through enforcing $m^{(i)}_{o_{\text{init}}} = 1$. Additionally, we initialize $\mu^{(i)}$'s from the templates bank $\mathcal{B} = \{\textbf{b}^{(i)}\}_{i=1}^{B} $
\begin{equation*}
    \scalebox{0.93}{$\displaystyle
            \mu_{i, j} =
            \begin{cases}
                +\kappa, & j \in \textbf{b}^{(i)}    \\
                -\kappa, & j \notin \textbf{b}^{(i)}
            \end{cases}
            \quad \text{for all } i \in [B] \text{ and } j \in [D].
        $}
\end{equation*}
where $\kappa \in \mathbb{R}_{>0}$. Note that one can use $\mathcal{B}\equiv\widehat{\mathcal{B}}^{(R)}$ in \autoref{alg:mutate_greedy_search} to warm start the continuous optimization.

\textbf{Differentiable Actor over Templates}\quad Given the current observation $(x_\textbf{o}, \textbf{o})$, we train an actor $\pi_{\theta}$ to output a categorical distribution over $B$ templates\footnote{We implement $\textbf{o} \subseteq \left[D\right]$ as a binary mask and $x_{\textbf{o}}$ as $x \odot \textbf{o}$. }:
\begin{equation*}
    \scalebox{0.90}{
        $h_{x_\textbf{o}} = f_\theta\left(\left[x_{\textbf{o}},\; \textbf{o}\right]\right) \in \mathbb{R}^B \quad \pi_\theta\left(x_\textbf{o},\textbf{o}\right)=\mathrm{softmax}(h_{x_o})\in\Delta^{B-1}$
    }
\end{equation*}
To select a template while remaining differentiable, we employ the straight-through Gumbel-Softmax technique \citep{jang2016categorical}. We use a single draw from the Gumbel-Softmax distribution $\widetilde{z} \sim \mathrm{GumbelSoftmax}(h_{x_{\textbf{o}}};\tau) \in \Delta^{B-1}$ during back-propagation for end-to-end training, where $\tau>0$ is a hyperparameter. To make a hard discrete choice during forward propagation, we select the template with the highest probability $z = \argmax{\widetilde{z}}$.

\textbf{Training Objective}\quad Using the template chosen with soft selection to select features for an instance is computed as $\overline{s} \triangleq \widetilde{M}\ \cdot \widetilde{z}$ (with hard analogue $s=Mz$ utilizing hard templates $M$). Thus, we can directly optimize the negative cumulative reward objective (\autoref{eq:subset_feature_loss_fn}) in terms of \emph{both the actor and the parameterized templates}:
\begin{equation}
    \scalebox{0.77}{$ \displaystyle
            \mathcal{L}_{\text{actor}}(\theta,\mu; x,y,\textbf{o}) = \ell\left( \widehat{y}_\phi\left(x\odot \textbf{o}^+, \textbf{o}^+\right), y\right) + \lambda \sum_{j=1}^D c(j)\left[ \overline{\textbf{o}}_j - \textbf{o}_j\right]_+
        $}
    \label{eq:actorloss}
\end{equation}
where $\widehat{y}_\phi$ is a (differentiable w.r.t. inputs) predictor, $\textbf{o}^+ = \textbf{o} + \overline{s}\odot (1-\textbf{o})$ represents the updated observed features after following the selected template and $[\overline{s}_j - \textbf{o}_j]_+ \triangleq \max\left( \overline{s}_j - \textbf{o}_j, 0\right)$ is introduced to capture the cost of newly acquired features.

\begin{algorithm}[t]
    \caption{\textsc{ContinuousTemplateRefinement}}
    \label{alg:training_procedure}
    \small
    \begin{algorithmic}[1]
        \REQUIRE training data~$\mathcal{D}_{\text{trn}}$, actor~$\pi_\theta$; template params~$\mu$, initial mask~$O_0$, initial feature~$o_\text{init}$, temperature~$\tau$, learning rate~$\eta$, batch size~$U$, number of features $D$.
        \ENSURE Refined $(\theta,\mu)$ and $\widetilde{\mathcal{B}}=\{\widetilde{\mathbf{b}}^{(i)}\}_{i=1}^B$
        \WHILE{training}
        \STATE $(X,Y)\sim\mathcal{D}_{\text{trn}}$;
        \FOR{$t=0,\dots,D-1$}
        \STATE \scalebox{0.95} {$\displaystyle(Z_t,\widetilde{Z}_t)\sim \textsc{GumbelSoftmax}\!\left(f_\theta\!\left([X_{O_t},\;O_t]\right);\tau\right)$}
        \STATE \scalebox{0.83}{$\displaystyle(\theta,\mu)\gets(\theta,\mu)-\eta\nabla_{\theta,\mu}\;\frac{1}{U}\sum_{n=1}^U \mathcal{L}_{\text{actor}}(\theta,\mu;X,Y,O_t)$} \COMMENT{\autoref{eq:actorloss}}
        \STATE $\mathcal{T}_t\gets \textsc{HardSelectTemplate}(Z_t,\mu)$
        \IF{$\mathcal{T}_t\setminus O_t=\varnothing$ }
        \STATE \textbf{break} \quad\COMMENT{stop if all batch instances are empty}
        \ENDIF
        \STATE $A_t\gets \textsc{SelectFeature}(\mathcal{T}_t\setminus O_t)$
        \STATE $O_{t+1}\gets O_t \lor \mathrm{OneHot}({A_t})$
        \ENDFOR
        \ENDWHILE
        \FOR{$i=1$ to $B$}
        \STATE $\widetilde{\mathbf{b}}^{(i)} \gets \{j\in[D]: \mu_{i,j} \ge 0\}\cup\{o_{\text{init}}\}$
        \ENDFOR
        \STATE \textbf{return} $(\theta, \mu), \widetilde{\mathcal{B}}$
    \end{algorithmic}
    \normalsize
\end{algorithm}
\textbf{Training Procedure}\quad Let $U$ be the batch size. We denote the batch tensors $X=[x^{(u)}]_{u=1}^U$, $Y=[y^{(u)}]_{u=1}^U$, and the observations at step $t$ be $O_t=[o_t^{(u)}]_{u=1}^{U}\in\{0,1\}^{U\times D}$. We initialize $O_0$ to all zeros except for the initial feature, i.e., the $o_{\text{init}}$-th column of $O_0$ is set to one. See \autoref{alg:training_procedure} for a pseudocode. We perform a $D$-step rollout: at each time step $t$, we sample hard/soft templates $(Z_t,\widetilde{Z}_t)$ via Gumbel-Softmax, update $(\theta,\mu)$ using $\mathcal{L}_{\text{actor}}$, and advance the current observation masks $O_{t+1}$ with the new feature $A_t$ until no new feature is needed.\looseness-1

\subsection{Interpretable Policy and Distillation}\label{sec:trees}
Rolling out our policies results in dynamic subsets of acquisitions that vary from instance to instance. Here, we wish to gain further insight into \emph{what subsets are acquired for what instances}, essentially distilling what (variable) information is important depending on instance characteristics. While scarcely explored in previous studies, such knowledge is often important in real-world problems; for example, in healthcare applications, an understanding of policy acquisitions distills variable informational-importance according to different patient sub-populations (e.g., younger underweight patients vs.~older obese patients). Here, we propose a knowledge-distillation strategy to yield interpretable policies and a breakdown of possible acquisitions by rules.

\textbf{Step-wise Student Decision Trees}\quad We collect state-action pairs from expert rollouts using an imitation framework (e.g., DAgger \citep{ross-2011}). We propose to distill a TAFA policy (e.g., \autoref{sec:template_based_policy} or \autoref{sec:gumbel}) into an intrinsically interpretable student policy by training a \textit{step-wise} ensemble of decision trees. Specifically, we maintain one tree per acquisition step $t$, denoted as $\{\text{Tree}_t\}_{t=1}^{K}$, where $K\leq D$ is the maximum number of acquisition steps considered in the rollouts. Training a separate $\text{Tree}_t$ per acquisition step decomposes the one difficult global decision problem into multiple simpler local sub-problems, allowing each tree to focus on the specific acquisition stage. The decision tree $\text{Tree}_t$ trains on teacher rollout observations $(x_\textbf{o}, \textbf{o})$'s where $|\textbf{o}|=t$ and outputs the discrete template index $v = \text{Tree}_t(x_\textbf{o}, \textbf{o}) \in \{1, ..., B\} \cup \{\varnothing\}$, indicating which template $\textbf{b}^{(v)}\in \mathcal{B}$ to follow next.

\begin{wrapfigure}[12]{R}{0.5\linewidth}
    \vspace{-2em}
    \centering
    \includegraphics[width=1.0\linewidth]{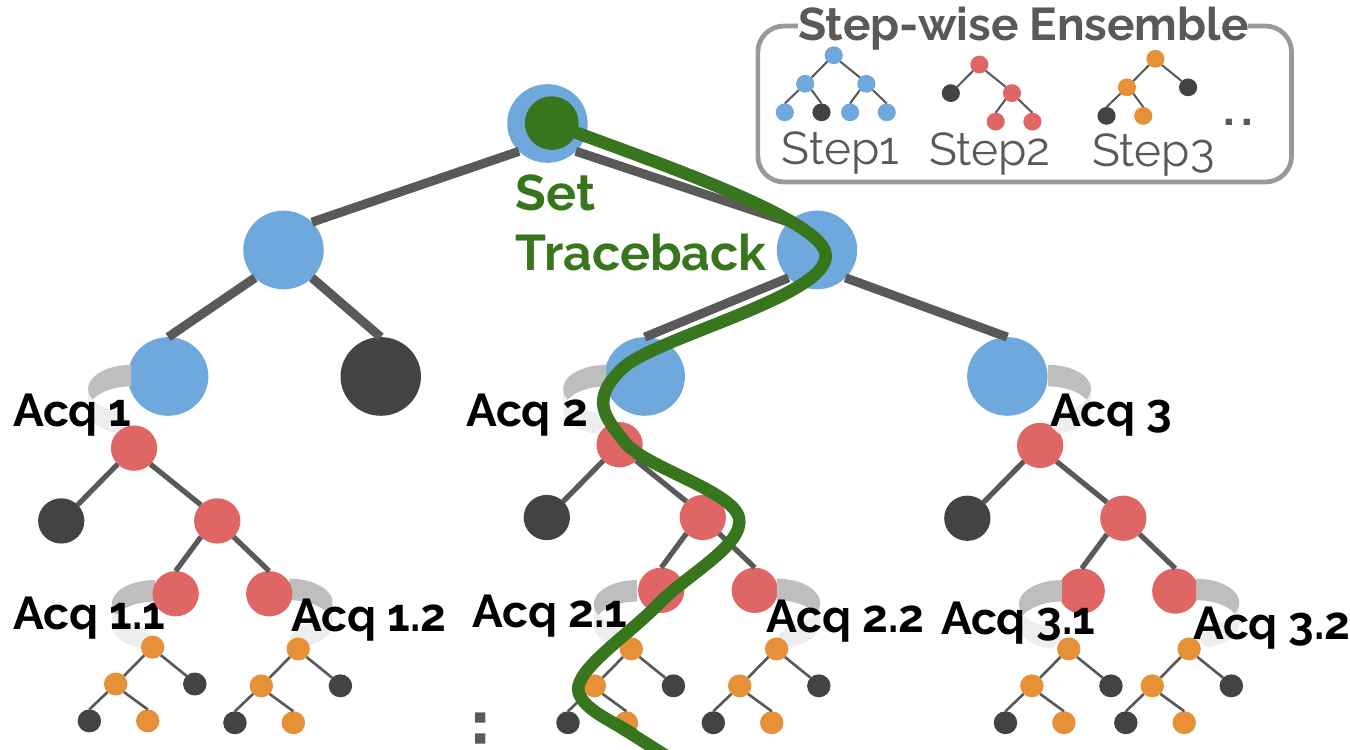}
    \caption{Step-wise Ensemble. At each step, non-termination (colored) leaves select features and advance to the next tree level, reusing the same tree structure to build complete acquisition paths that form interpretable rules.}
    \label{fig:extract-rule-fig}
\end{wrapfigure}
\textbf{Tree Stitching for Subset/Rule(s) Extraction}\quad At the leaves of $\text{Tree}_t$, the policy shall either: 1) terminate, yielding the acquired subset with a traceback; or 2) select a template, with a corresponding feature to acquire (using feature selection step; \autoref{eq:feature_acquisition}). When not terminating, the policy continues by un-rolling $\text{Tree}_{t+1}$ with the updated current set $|o| = t+1$. Thus, the ensemble's policy can be represented as a single large stitched tree (re-using the appropriate $\text{Tree}_{t}$ at levels), where all leaves terminate acquisition with a specified subset (see \autoref{fig:extract-rule-fig}).

Hence, one may collect all possible acquisitions by aggregating the unique subsets found at the leaves of the stitched tree. Moreover, for each unique subset, one can perform a traceback to collect one or more (if the acquired subset occurred at multiple leaves) rules that determine the respective acquisition set, i.e., \emph{acquire} $\textit{set}_1\!\!: \left\{ x_6, x_7, x_9 \right\}$ \texttt{if} $\mathrm{rule}_{1.1}\!\!: \ 0.0<x_6\leq 1.0$
\texttt{or} $\mathrm{rule}_{1.2}\!\!: \ldots$; \emph{acquire} $\textit{set}_2\!\!: \left\{ x_1, x_4, x_6 \right\}$ \texttt{if} $\mathrm{rule}_{2.1}\!\!: x_6\leq 0.0 \, \wedge x_1\leq 0.0$; etc. Note that sets and rules can be ordered by training set frequency for further distillation.

\section{Experiments}
We evaluate different variants of our proposed TAFA method against existing baselines across three key dimensions: prediction accuracy, feature acquisition efficiency, and computational runtime. Specifically, the TAFA variants we considered are as follows: 1) \texttt{tafa-knn} Gumbel templates (\autoref{sec:gumbel}) with k-NN policy (\autoref{eq:template_selection}, \autoref{sec:template_based_policy}); 2) \texttt{tafa-actor} Gumbel templates with Gumbel policy (\autoref{sec:gumbel}); 3) \texttt{tafa-interp} distilled step-wise decision tree policy (\autoref{sec:trees}) with mutation-searched templates (\autoref{sec:mutate_search}). Code will be released upon publication.

Experiments span one synthetic \texttt{cube}-$\sigma = 0.3$~\citep[$D=20$;][]{ruec-2012} and five real-world datasets: the Big Five Personality questionnaire data\footnote{\url{https://openpsychometrics.org/_rawdata/}} (\texttt{big5}, $D=50$),  Gas sensor home activity monitoring~\citep[\texttt{gas}, $D=10$;][]{huer-2016}, Electrical Grid stability data~\citep[\texttt{grid}, $D=11$;][]{arza-2018}, MNIST~\citep[\texttt{mnist}, $D=256$, $16 \times 16$ pxs;][]{deng-2012} and Fashion MNIST~\citep[\texttt{fashion}, $D=256$, $16 \times 16$ pxs;][]{xiao-2017}.

We use uniform acquisition costs $c(d) = 1$ for all features $d \in [D]$ and negative cross-entropy as the task objective; the parameter $\lambda$ in \autoref{eq:subset_feature_loss_fn} was varied to control the trade-off between prediction accuracy and acquisition cost. Unless otherwise noted, we use a neural network as the arbitrary subset feature predictor $\widehat{y}\left(x_{\textbf{o}}\right)$. \autoref{sec:implementation} has details about choice of $\lambda$ and neural network architecture.

\subsection{Benchmark Dataset Evaluations}
We compare against SOTA RL-free AFA methods---AACO~\citep[\texttt{aaco};][]{vala-2024}, SEFA~\citep[\texttt{SEFA};][]{norcliffe2025stochastic}, DIME~\citep[\texttt{dime};][]{gadg-2024}---as well as a static acquisition baseline (always acquiring the same feature subset sequentially, \texttt{static}). Importantly, TAFA is also RL-free, making these comparisons methodologically aligned; moreover, each of the cited approaches is shown in its respective work to match or outperform RL-based acquisition policies. We conduct additional comparisons to interpretable RL policies below (\autoref{sec:interp-analysis}).

We emphasize that, due to the difficulty of active feature acquisition in high-dimensional spaces, prior AFA studies typically operate in low-dimensional settings (e.g., SEFA~\citep{norcliffe2025stochastic} acquires among 20 prechosen MNIST features). In contrast, we retain a high-dimensional MNIST setting (up to 256 acquisitions) and evaluate acquisition over hundreds of candidate features without such preselection.

\begin{figure}[h]
    \centering
    \includegraphics[width=1.0\linewidth]{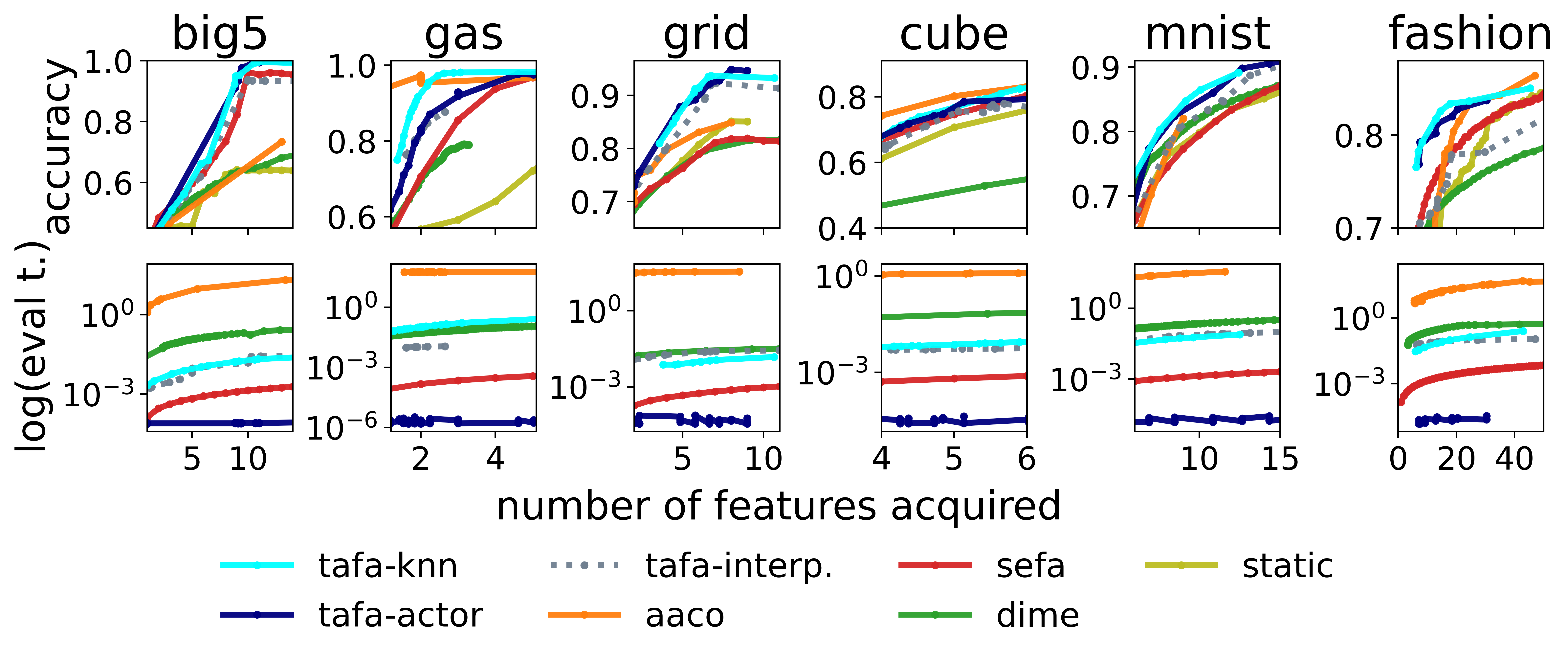}
    \caption{Accuracy vs. \emph{log-scaled} inference time vs. number of features acquired compared to baselines. Tick marks for methods are computed by adjusting respective cost/benefit trade-off hparams.}
    \label{fig:acc_vs_nfeats}
\end{figure}

The first row of \autoref{fig:acc_vs_nfeats} shows that TAFA consistently outperforms baselines across most datasets, achieving higher accuracy with fewer feature acquisitions. The second row reports the average inference time per policy, plotted on a logarithmic scale to account for orders-of-magnitude differences. TAFA is substantially faster than competing methods in most settings---particularly relative to the other cost/tradeoff based \texttt{aaco} approach---highlighting its computational efficiency for time-sensitive and real-time applications. Among TAFA variants, the fully differentiable \texttt{tafa-actor} is notably faster than \texttt{tafa-knn} and \texttt{tafa-interp}, reflecting the benefits of amortized decision-making.

\begin{figure}[h]
    \centering
    \includegraphics[trim={0cm 0cm 0cm 0cm}, clip, width=.9\linewidth]{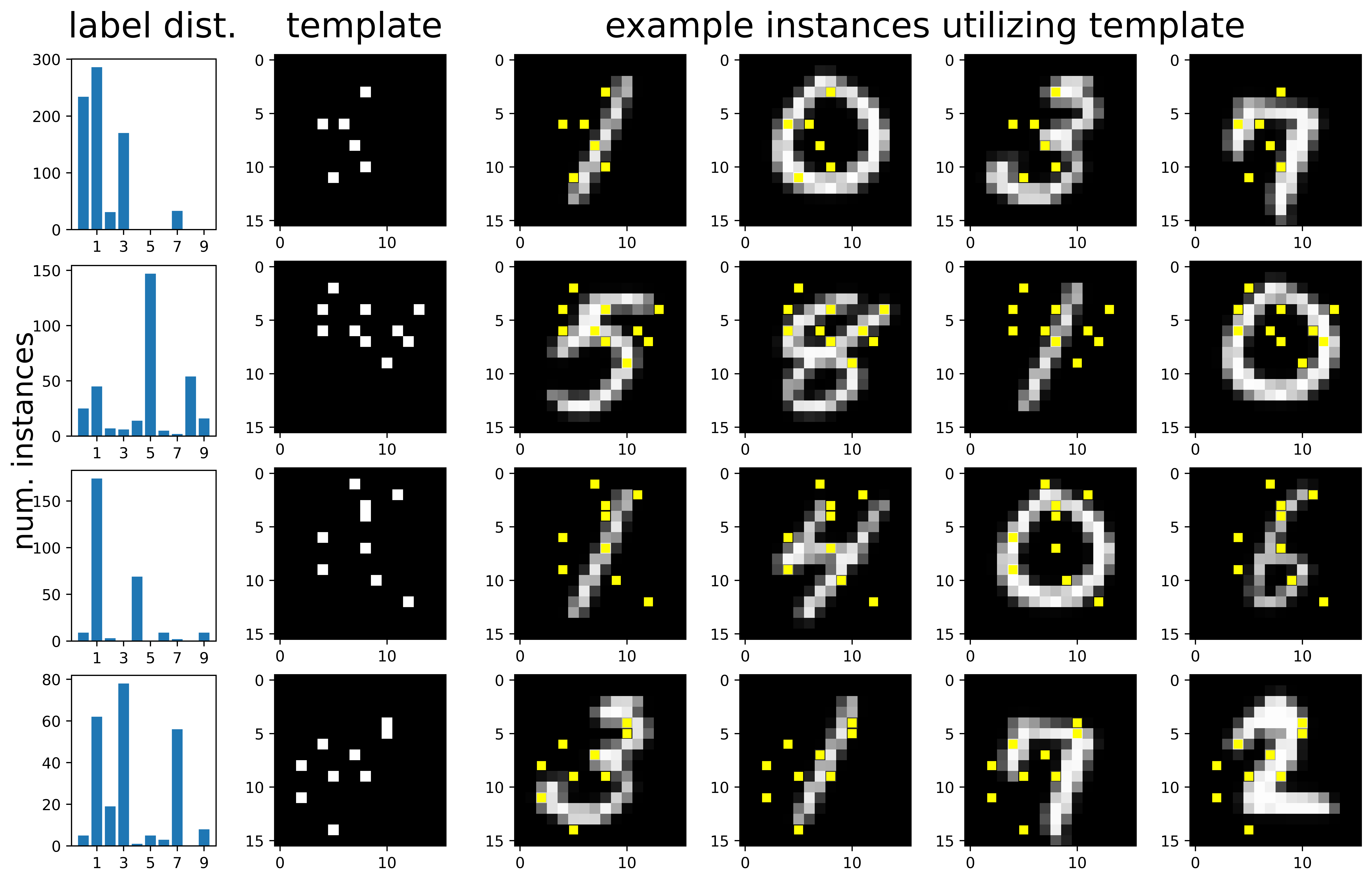}
    \caption{Visualization of feature templates found for the MNIST dataset. Yellow pixels in the ``example instances utilizing templates'' column represent the pixels that are acquired by the TAFA agent to classify the digits of interest.}
    \label{fig:mnist-templates}
    \vspace{1em}
    \centering
    \includegraphics[width=1.0\linewidth]{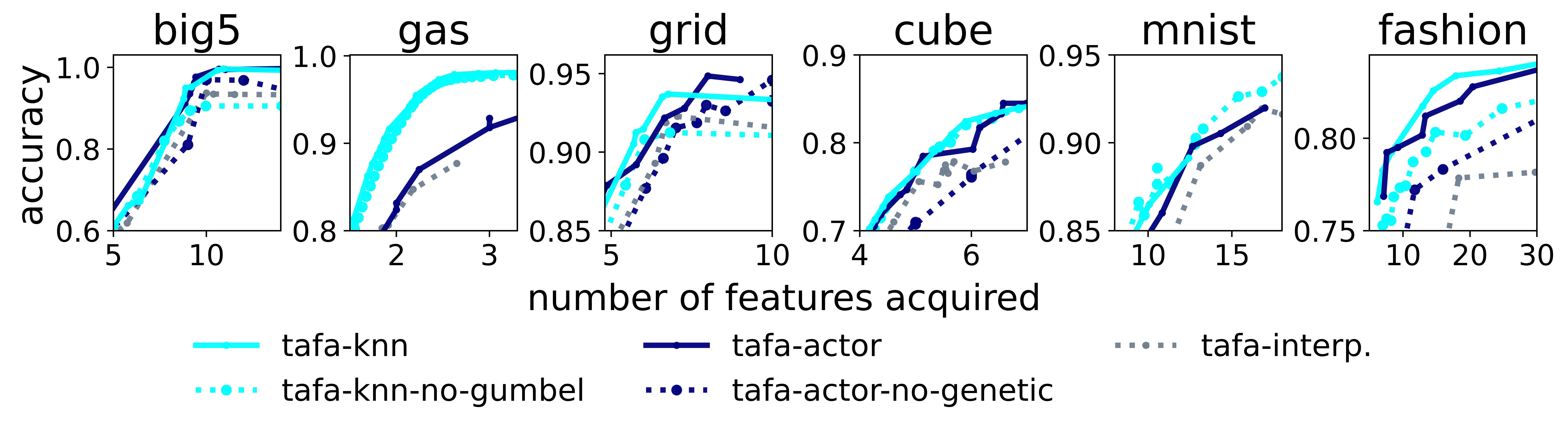}\\
    \begin{subfigure}{0.62\linewidth}
        \centering
        \includegraphics[width=0.93\linewidth]{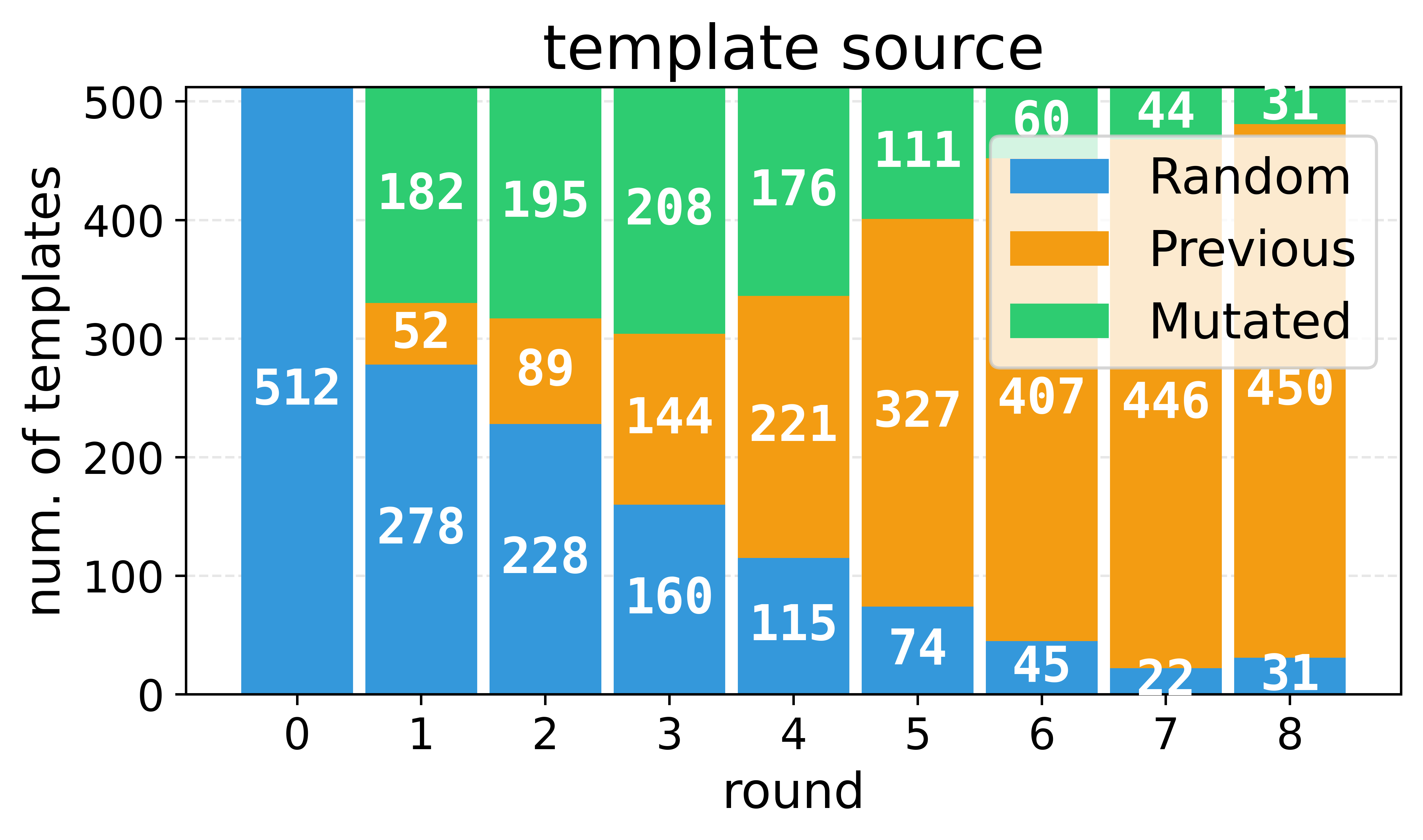}
    \end{subfigure}
    \begin{subfigure}{0.37\linewidth}
        \centering
        \includegraphics[width=0.93\linewidth]{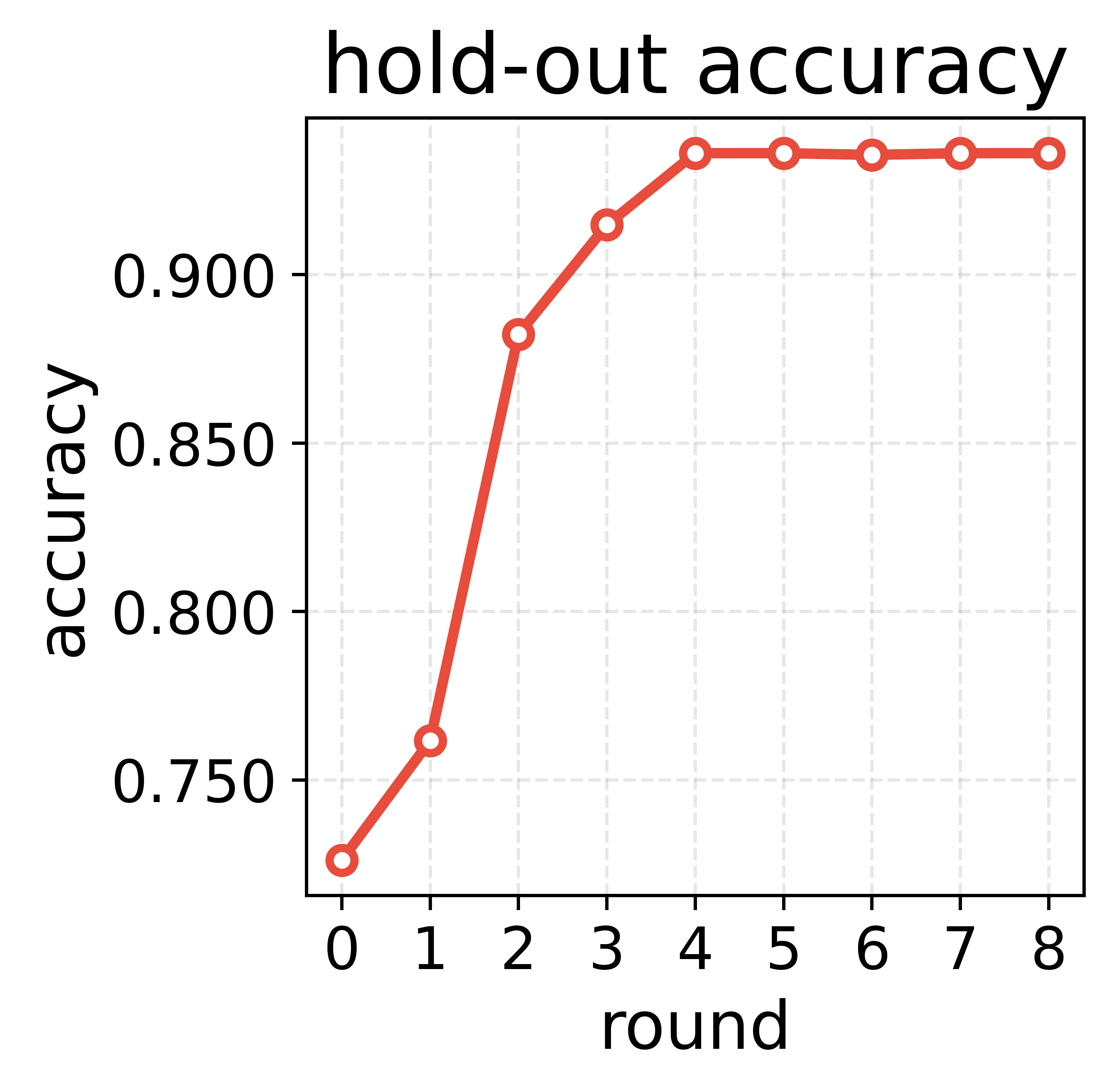}
    \end{subfigure}
    \caption{(Top) Accuracy vs. number of features acquired. (Bottom) Template composition (left) and test time performance (right) over the course of mutation on \texttt{big5} dataset.}
    \label{fig:mutation-progres}
\end{figure}
\textbf{Learned Templates}\quad We note that all TAFA variants expose some level of interpretability through an analysis of the learned templates.
For example, when visualizing templates discovered for MNIST and corresponding instances mapped to each template (\autoref{fig:mnist-templates}), one can see that TAFA learns to focus on the distinctive curves and edges that characterize different digits.
This direct interpretability is in contrast to prior work, which does not expose any summary of canonical subsets; below, we further analyze our interpretable policies.

\subsection{TAFA Variants Ablation}
In \autoref{fig:mutation-progres}, we validate our method through a study of multiple TAFA variants.
\begin{enumerate}[leftmargin=*,topsep=0ex, nosep, noitemsep]
    \item \textbf{Our continuous-relaxation optimization often helps find better templates.} We see that in several datasets (e.g., \texttt{big5}, \texttt{grid}, \texttt{fashion}) the performance of a policy using templates fine-tuned with the continuous relaxation  (\texttt{tafa-knn}, \autoref{sec:gumbel}) improves over the same policy with the templates before the relaxation tuning (\texttt{tafa-knn-no-gumbel}).
    \item \textbf{Genetic optimization often provides a helpful initialization to continuous relaxation optimization of templates.} We see that in several datasets (e.g., \texttt{big5}, \texttt{grid}, \texttt{cube}, \texttt{fashion}),  continuous relaxation optimization over randomly initialized templates (\texttt{tafa-actor-no-genetic}) is not as effective as when fine-tuning the genetic templates (\texttt{tafa-actor}, \autoref{sec:mutate_search}).
    \item \textbf{Our interpretable distillation retains performance.} We see relatively small degradation in accuracy/cost trade-offs utilizing our distillation techniques (\texttt{tafa-interp}, \autoref{sec:trees}) despite the added interpretability of simple rules (see \autoref{sec:interp-analysis}).
    \item \textbf{Our iterative genetic approach converges rapidly.} We visualize the genetic optimization procedure (\autoref{alg:mutate_greedy_search}) in \autoref{fig:mutation-progres} for the \texttt{big5} dataset as a function of rounds. The resulting policy quickly converges in terms of average accuracy (\autoref{fig:mutation-progres}, right). Moreover, by visualizing the source of the template set at each round (\texttt{random}, a randomly generated template; \texttt{mutated}, a mutated template from the last round's optimized template collection; \texttt{previous} a direct template from the last round's optimized template collection), we see that our mutations are providing useful alternatives in early rounds, and the optimized set converges relatively quickly after (choosing to preserve previous templates).
\end{enumerate}
Observations 1 and 2 show that our optimization techniques are complementary and pair well together; observation 3 shows the effectiveness of our interpretable policies; observation 4 further validates the design of our optimization.

\subsection{Interpretability Analysis}\label{sec:interp-analysis}
We compare our interpretable policy against established interpretable RL baselines, conduct ablative studies to validate our design choices, and provide qualitative analysis on synthetic and real-world datasets; \autoref{sec:interp-appx} has a further qualitative study around additional automotive diagnosis.

\begin{figure}[b]
    \centering
    \begin{subfigure}{1.0\linewidth}
        \includegraphics[width=\linewidth]{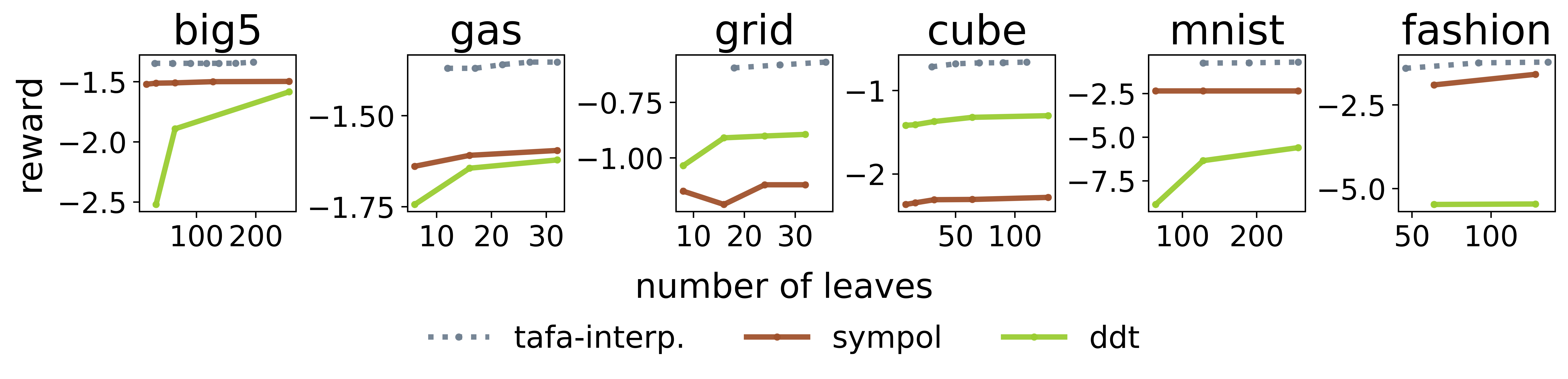}
    \end{subfigure}
    \hfill
    \begin{subfigure}{1.0\linewidth}
        \includegraphics[width=\linewidth]{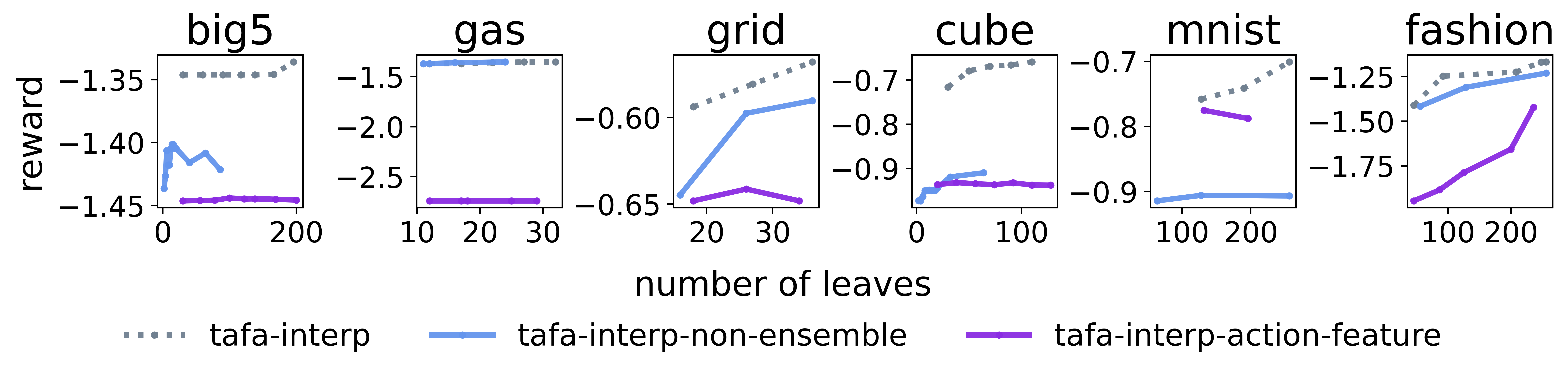}
    \end{subfigure}
    \caption{Reward vs. number of leaves acquired. (Top) Comparing to \emph{baselines}. (Bottom) Comparing to \texttt{tafa-interp} variants. A higher reward is better for comparable leaf budgets.}
    \label{fig:interp-reward-leaves-ablation}
\end{figure}
\textbf{Interpretable Baseline Comparisons}\quad We compare our interpretable policy (\texttt{tafa-interp}) to two interpretable RL baselines: 1) differentiable decision tree ~\citep[\texttt{ddt}][]{silva2020ddt} and 2) symbolic tree-based on-policy RL ~\citep[\texttt{sympol}][]{marton2024sympol}. DDT trains differentiable decision trees with PPO \citep{schulman2017ppo}, then post-hoc discretizes them, while SYMPOL directly optimizes the hard decision tree with PPO, avoiding discretization loss.
Following prior work, we treat feature indices as actions and train with PPO to support non-greedy acquisition. For consistent results, \texttt{tafa-interp}, \texttt{ddt}, and \texttt{sympol} are trained with the same classifier and fixed $\lambda$; see \autoref{sec:implementation} for details. Across datasets, \autoref{fig:interp-reward-leaves-ablation} (top) shows \texttt{tafa-interp} consistently achieves the highest reward at comparable leaf budgets \citep{roth2019conservative, glanois2024survey} to \texttt{ddt} and \texttt{sympol}, indicating that \texttt{tafa-interp} can make better performance-cost-aware decisions under similar interpretability constraints compared to the baselines.

\textbf{Ablative Studies}\quad Two variants of \texttt{tafa-interp} are trained; \texttt{tafa-interp-action-feature}: the decision tree predicts feature indices (\autoref{eq:feature_acquisition}) instead of template indices; \texttt{tafa-interp-non-ensemble}: a single tree is shared across cardinalities. \autoref{fig:interp-reward-leaves-ablation} (bottom) shows that \texttt{tafa-interp} achieves higher reward at comparable leaf budgets than its variants, indicating (1) acting on template action space and (2) using step-wise trees are both beneficial for optimizing the tree policy under the reward objective.

\begin{wrapfigure}[9]{R}{0.45\linewidth}
    \centering
    \includegraphics[width=1.0\linewidth, trim={0cm 2cm 7cm 0cm}, clip]{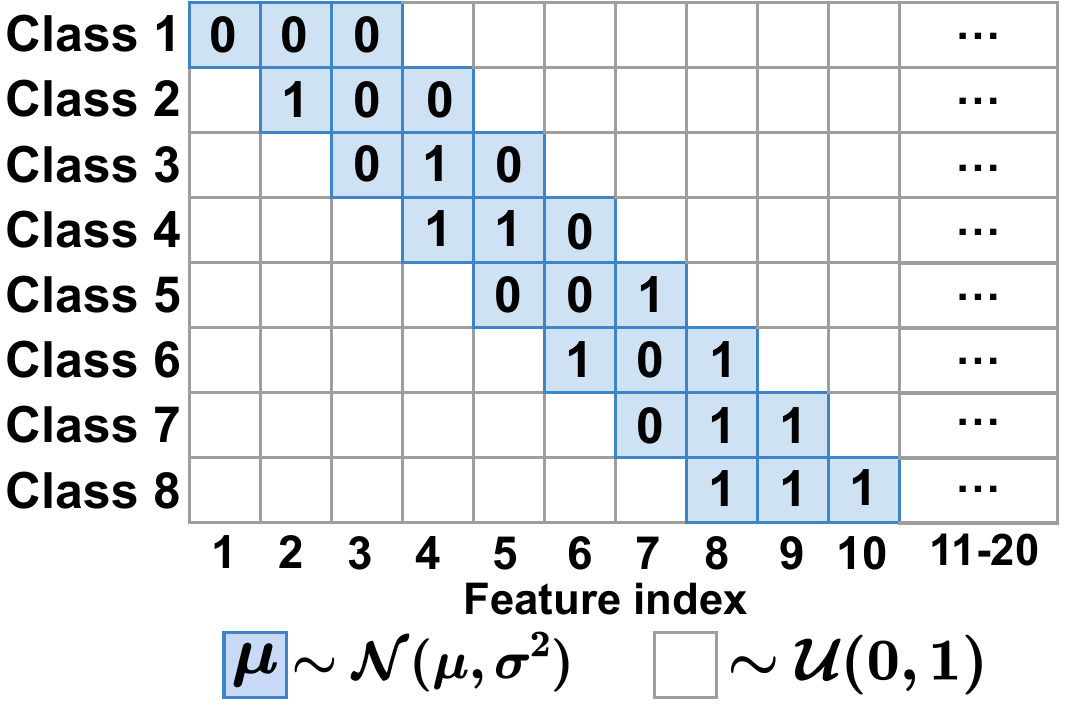}
    \caption{Distribution of $\texttt{CUBE-}\sigma$ dataset.}
    \label{fig:cube_desc.pdf}
\end{wrapfigure}
\textbf{Qualitative Analysis on $\texttt{CUBE-}\sigma$ dataset}\quad We use the $\texttt{CUBE-}\sigma=0.3$ dataset \cite{shim-2018}, a synthetic dataset with $D=20$ features and $8$ classes where each instance contains 17 uniform noise features and 3 informative Gaussian features. For class $y=c$, only the contiguous block $(c, c+1, c+2)$ is normally distributed (class distribution in \autoref{fig:cube_desc.pdf}, Gaussian means in blue cells), while all the remaining features are uniform noise.

\begin{figure}[h]
    \centering
    \includegraphics[width=.95\linewidth]{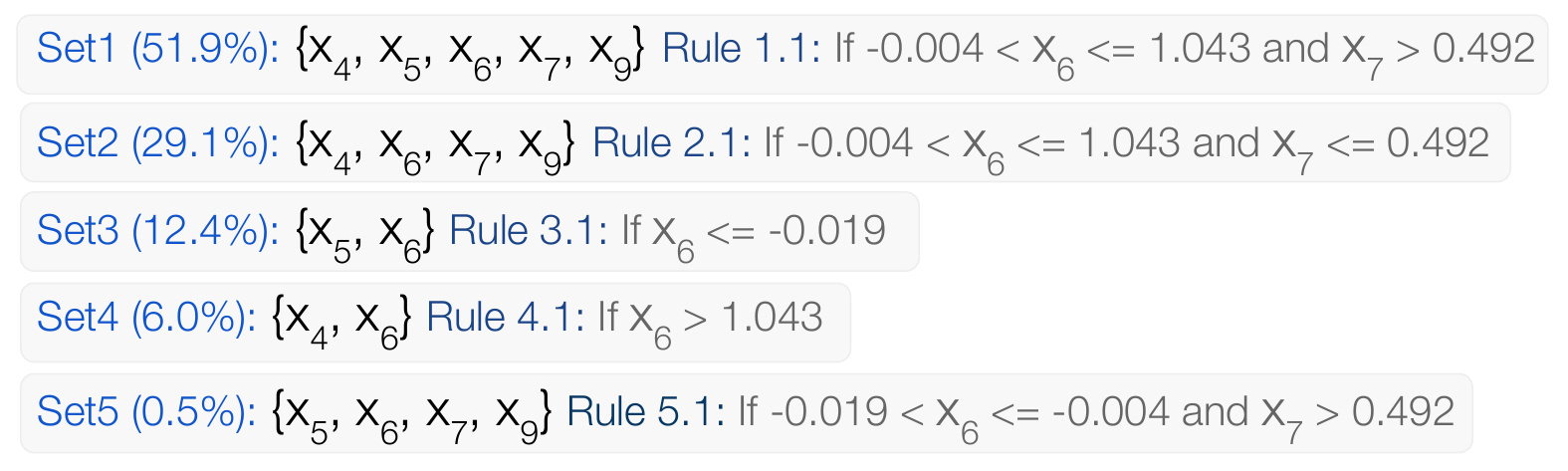}
    \caption{Extracted rules on the CUBE dataset.}
    \label{fig:posthoc_cube}
\end{figure}
\autoref{fig:posthoc_cube} shows an example of the extracted subsets of acquisitions and the respective rules (\autoref{sec:trees}). First, we note the efficacy of our method to very succinctly distill down the behavior of the policy for nearly all instances (99.9\%) into 5 distinct subsets. This is in stark contrast to previous works' black-box policies. Second, we can confirm that the learned policy is behaving reasonably based on a quick analysis of the subsets/rules. For example, \texttt{Set3} is selected if \uline{$x_6$ is negative} ($x_6<-0.019$), which rules out any class where $x_6$ is uniform ($y \notin \{1, 2, 3, 7, 8\}$, with the negative value indicating that a $\mathcal{N}(0,\sigma^2)$ class for $x_6$ is more likely ($y\in\{4,5\}$), and correspondingly acquires $x_5$ to resolve the ambiguity. \texttt{Set2} is selected based on checking if \uline{$x_6$ is likely uniform} ($-0.004<x_6\leq 1.043$) and \uline{$x_7$ is smaller} ($x_7\leq 0.492$, more likely $\mathcal{N}(0,\sigma^2)$ or potentially uniform), indicating that the class may be $y\in\{1,2,3,7,8\}$, and correspondingly acquires $x_4$ and $x_9$ (resolving possibilities of classes $1,2,3$ and $7,8$, respectively). \texttt{Set1} corresponds to the previous case, but when \uline{$x_7$ is large} (likely either $\mathcal{N}(1,\sigma^2)$ or potentially uniform), $y\in\{1,2,3,5,8\}$; thus, the policy additionally acquires $x_5$ to resolve class ambiguity\looseness-1.

\begin{figure}[h]
    \centering
    \includegraphics[width=.95\linewidth, trim={0cm 6.5cm 0cm 0cm}, clip]{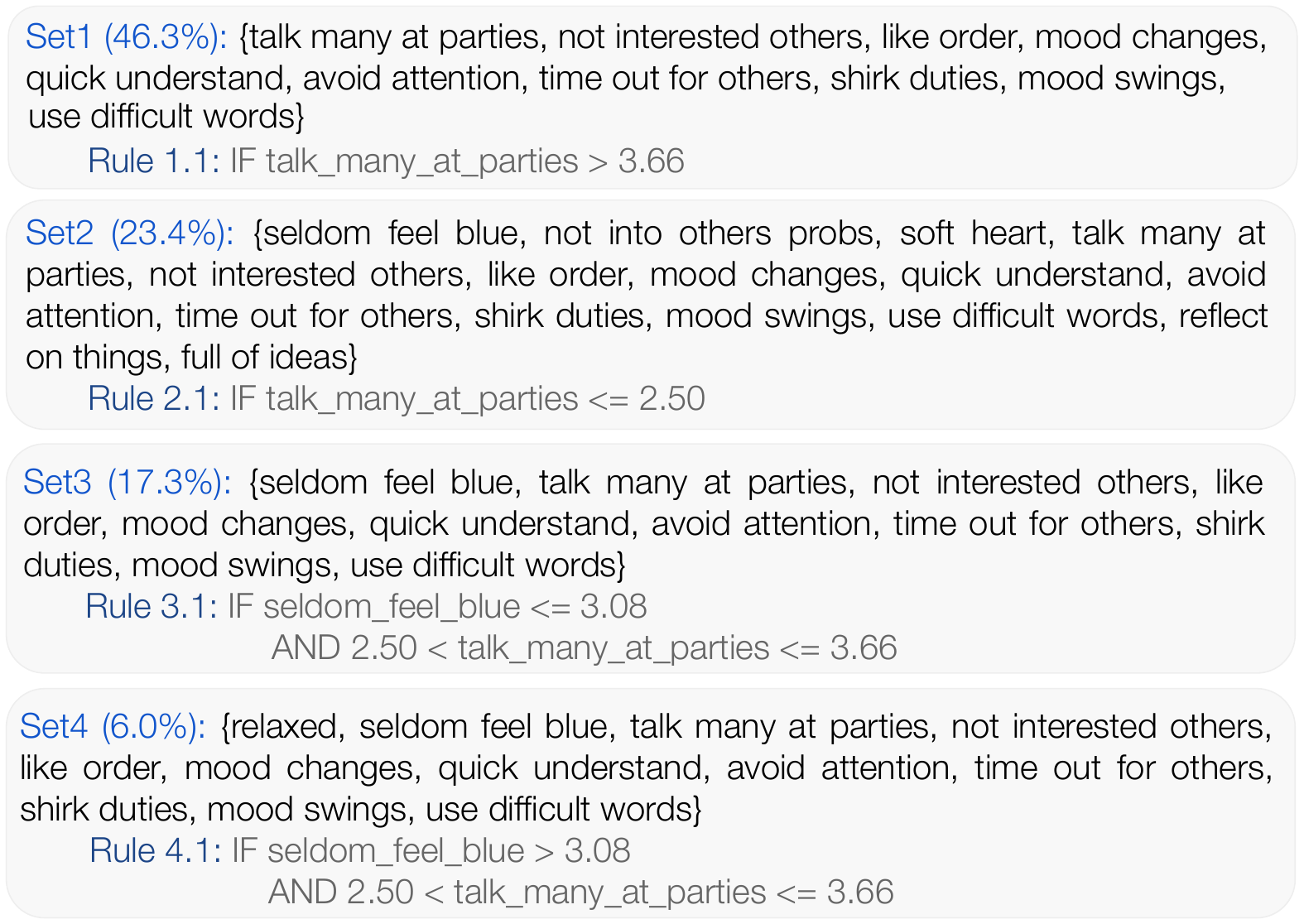}
    \caption{Extracted rules on the Big5 dataset.}
    \label{fig:posthoc_big5}
\end{figure}

\textbf{Psychological Analysis}\quad We use the Big Five Personality questionnaire data (\texttt{big5}). Each instance contains 50 survey questions rated on a 5-point agreement scale. Following \citet{vala-2024}, the label is a 4-class target (0-3) corresponding to Emotional Stability (ES) score quartiles. We see that our method is again able to succinctly distill a policy to a handful of rules, acquiring informative features to emotional stability (\autoref{fig:posthoc_big5}).

\section{Conclusion}
This work addresses a previously underexplored question in AFA: \emph{what are the canonical feature subsets driving the cost/benefit performance?} We answer this by introducing template-based active feature acquisition (TAFA), a framework that organizes acquisition decisions around a learned dictionary of informative feature subsets, improving both effectiveness and inference efficiency. We formulate template discovery as a set optimization problem and show that the resulting objective is submodular, yielding theoretical approximation guarantees. Building on this structure, we develop an iterative genetic search procedure with continuous relaxation-based refinement. For interpretability, we distill TAFA into an ensemble of step-wise decision trees operating on templates rather than individual features, producing shorter and more transparent decision paths than existing interpretable RL-based approaches. Across synthetic and real-world benchmarks, TAFA consistently outperforms state-of-the-art baselines while achieving lower acquisition cost and substantially reduced computation, highlighting its practicality, scalability, and interpretability in real-world—and potentially real-time—AFA settings.

\section*{Impact Statement}
This paper presents work whose goal is to advance the field of Machine
Learning. There are many potential societal consequences of our work, none
which we feel must be specifically highlighted here.

\bibliography{refs}

@InProceedings{vala-2024,
  author       = {Valancius, Michael and Lennon, Maxwell and Oliva, Junier},
  booktitle    = {International Conference on Machine Learning},
  title        = {Acquisition Conditioned Oracle for Nongreedy Active Feature Acquisition},
  year         = {2024},
  organization = {PMLR},
  pages        = {48957--48975},
}

@Article{forr-1996,
  author    = {Forrest, Stephanie},
  journal   = {ACM computing surveys (CSUR)},
  number    = {1},
  pages     = {77--80},
  title     = {Genetic algorithms},
  volume    = {28},
  year      = {1996},
  publisher = {ACM New York, NY, USA},
}

@Book{siva-2008,
  author    = {Sivanandam, SN and Deepa, SN and Sivanandam, SN and Deepa, SN},
  publisher = {Springer},
  title     = {Genetic algorithms},
  year      = {2008},
}

@Article{holl-1973,
  author    = {Holland, John H},
  journal   = {SIAM journal on computing},
  number    = {2},
  pages     = {88--105},
  title     = {Genetic algorithms and the optimal allocation of trials},
  volume    = {2},
  year      = {1973},
  publisher = {SIAM},
}

@InProceedings{ross-2011,
  author       = {Ross, St{\'e}phane and Gordon, Geoffrey and Bagnell, Drew},
  booktitle    = {Proceedings of the fourteenth international conference on artificial intelligence and statistics},
  title        = {A reduction of imitation learning and structured prediction to no-regret online learning},
  year         = {2011},
  organization = {JMLR Workshop and Conference Proceedings},
  pages        = {627--635},
}

@Article{saar-2009,
  author    = {Saar-Tsechansky, Maytal and Melville, Prem and Provost, Foster},
  journal   = {Management Science},
  number    = {4},
  pages     = {664--684},
  title     = {Active feature-value acquisition},
  volume    = {55},
  year      = {2009},
  publisher = {INFORMS},
}

@Article{rahb-2025,
  author  = {Rahbar, Arman and Aronsson, Linus and Chehreghani, Morteza Haghir},
  journal = {arXiv preprint arXiv:2502.11067},
  title   = {A Survey on Active Feature Acquisition Strategies},
  year    = {2025},
}

@InProceedings{dula-2011,
  author       = {Dulac-Arnold, Gabriel and Denoyer, Ludovic and Preux, Philippe and Gallinari, Patrick},
  booktitle    = {Machine Learning and Knowledge Discovery in Databases: European Conference, ECML PKDD 2011, Athens, Greece, September 5-9, 2011. Proceedings, Part I 11},
  title        = {Datum-wise classification: a sequential approach to sparsity},
  year         = {2011},
  organization = {Springer},
  pages        = {375--390},
}

@Article{shim-2018,
  author  = {Shim, Hajin and Hwang, Sung Ju and Yang, Eunho},
  journal = {Advances in neural information processing systems},
  title   = {Joint active feature acquisition and classification with variable-size set encoding},
  volume  = {31},
  year    = {2018},
}

@InProceedings{li-2021,
  author       = {Li, Yang and Oliva, Junier},
  booktitle    = {International conference on machine learning},
  title        = {Active feature acquisition with generative surrogate models},
  year         = {2021},
  organization = {PMLR},
  pages        = {6450--6459},
}

@InProceedings{ma-2019,
  author       = {Ma, Chao and Tschiatschek, Sebastian and Palla, Konstantina and Hernandez-Lobato, Jose Miguel and Nowozin, Sebastian and Zhang, Cheng},
  booktitle    = {International Conference on Machine Learning},
  title        = {EDDI: Efficient Dynamic Discovery of High-Value Information with Partial VAE},
  year         = {2019},
  organization = {PMLR},
  pages        = {4234--4243},
}

@InProceedings{gadg-2024,
  author    = {Soham Gadgil and Ian Connick Covert and Su-In Lee},
  booktitle = {The Twelfth International Conference on Learning Representations},
  title     = {Estimating Conditional Mutual Information for Dynamic Feature Selection},
  year      = {2024},
  url       = {https://openreview.net/forum?id=Oju2Qu9jvn},
}

@Article{ding-2020,
  author    = {Ding, Zihan and Dong, Hao},
  journal   = {Deep Reinforcement Learning: Fundamentals, Research and Applications},
  pages     = {249--272},
  title     = {Challenges of reinforcement learning},
  year      = {2020},
  publisher = {Springer},
}

@Misc{arza-2018,
  author       = {Arzamasov, Vadim},
  howpublished = {UCI Machine Learning Repository},
  note         = {{DOI}: https://doi.org/10.24432/C5PG66},
  title        = {{Electrical Grid Stability Simulated Data }},
  year         = {2018},
}

@Misc{huer-2016,
  author       = {Huerta, Flavia and Huerta, Ramon},
  howpublished = {UCI Machine Learning Repository},
  note         = {{DOI}: https://doi.org/10.24432/C5BS4F},
  title        = {{Gas sensors for home activity monitoring}},
  year         = {2016},
}

@Article{deng-2012,
  author    = {Deng, Li},
  journal   = {IEEE Signal Processing Magazine},
  number    = {6},
  pages     = {141--142},
  title     = {The mnist database of handwritten digit images for machine learning research},
  volume    = {29},
  year      = {2012},
  publisher = {IEEE},
}

@Article{ruec-2012,
  author    = {Rückstieß, Thomas and Osendorfer, Christian and van der Smagt, Patrick},
  doi       = {10.1007/s13042-012-0092-x},
  journal   = {International Journal of Machine Learning and Cybernetics},
  number    = {3},
  pages     = {235--243},
  title     = {Minimizing data consumption with sequential online feature selection},
  volume    = {4},
  year      = {2012},
  issn      = {1868-808X},
  month     = apr,
  publisher = {Springer Science and Business Media LLC},
}

@Article{krau-2014,
  author = {Krause, Andreas and Golovin, Daniel},
  title  = {Submodular function maximization.},
  year   = {2014},
}

@InProceedings{krau-2008,
  author    = {Krause, Andreas and Horvitz, Eric},
  booktitle = {AAAI},
  title     = {A Utility-Theoretic Approach to Privacy and Personalization.},
  year      = {2008},
  pages     = {1181--1188},
  volume    = {8},
}

@Article{nemh-1978,
  author    = {Nemhauser, G. L. and Wolsey, L. A.},
  doi       = {10.1287/moor.3.3.177},
  journal   = {Mathematics of Operations Research},
  number    = {3},
  pages     = {177--188},
  title     = {Best Algorithms for Approximating the Maximum of a Submodular Set Function},
  volume    = {3},
  year      = {1978},
  issn      = {1526-5471},
  month     = aug,
  publisher = {Institute for Operations Research and the Management Sciences (INFORMS)},
}

@Article{Frieze1974,
  author    = {Frieze, Alan M},
  journal   = {Mathematical Programming},
  number    = {1},
  pages     = {245--248},
  title     = {A cost function property for plant location problems},
  volume    = {7},
  year      = {1974},
  publisher = {Springer},
}

@InProceedings{marton2024sympol,
  author    = {Marton, Sascha and Grams, Tim and Vogt, Florian and L{\"u}dtke, Stefan and Bartelt, Christian and Stuckenschmidt, Heiner},
  booktitle = {The Thirteenth International Conference on Learning Representations},
  title     = {Mitigating Information Loss in Tree-Based Reinforcement Learning via Direct Optimization},
  year      = {2025},
}

@InProceedings{silva2020ddt,
  author       = {Silva, Andrew and Gombolay, Matthew and Killian, Taylor and Jimenez, Ivan and Son, Sung-Hyun},
  booktitle    = {International conference on artificial intelligence and statistics},
  title        = {Optimization methods for interpretable differentiable decision trees applied to reinforcement learning},
  year         = {2020},
  organization = {PMLR},
  pages        = {1855--1865},
}

@Article{schulman2017ppo,
  author  = {Schulman, John and Wolski, Filip and Dhariwal, Prafulla and Radford, Alec and Klimov, Oleg},
  journal = {arXiv preprint arXiv:1707.06347},
  title   = {Proximal policy optimization algorithms},
  year    = {2017},
}

@Article{roth2019conservative,
  author  = {Roth, Aaron M and Topin, Nicholay and Jamshidi, Pooyan and Veloso, Manuela},
  journal = {arXiv preprint arXiv:1907.01180},
  title   = {Conservative q-improvement: Reinforcement learning for an interpretable decision-tree policy},
  year    = {2019},
}

@Article{glanois2024survey,
  author    = {Glanois, Claire and Weng, Paul and Zimmer, Matthieu and Li, Dong and Yang, Tianpei and Hao, Jianye and Liu, Wulong},
  journal   = {Machine Learning},
  number    = {8},
  pages     = {5847--5890},
  title     = {A survey on interpretable reinforcement learning},
  volume    = {113},
  year      = {2024},
  publisher = {Springer},
}

@Article{guney2025active,
  author  = {Guney, Osman Berke and Saichandran, Ketan Suhaas and Elzokm, Karim and Zhang, Ziming and Kolachalama, Vijaya B},
  journal = {Proceedings of machine learning research},
  pages   = {20748},
  title   = {Active feature acquisition via explainability-driven ranking},
  volume  = {267},
  year    = {2025},
}

@Article{lundberg2017unified,
  author  = {Lundberg, Scott M and Lee, Su-In},
  journal = {Advances in neural information processing systems},
  title   = {A unified approach to interpreting model predictions},
  volume  = {30},
  year    = {2017},
}

@Article{verg-2023,
  author    = {Vergara, Mary and Ramos, Leo and Rivera-Campoverde, Néstor Diego and Rivas-Echeverría, Francklin},
  doi       = {10.1109/access.2023.3331316},
  journal   = {IEEE Access},
  pages     = {126155--126171},
  title     = {EngineFaultDB: A Novel Dataset for Automotive Engine Fault Classification and Baseline Results},
  volume    = {11},
  year      = {2023},
  issn      = {2169-3536},
  publisher = {Institute of Electrical and Electronics Engineers (IEEE)},
}

@Misc{xiao-2017,
  author    = {Xiao, Han and Rasul, Kashif and Vollgraf, Roland},
  title     = {Fashion-MNIST: a Novel Image Dataset for Benchmarking Machine Learning Algorithms},
  year      = {2017},
  copyright = {arXiv.org perpetual, non-exclusive license},
  doi       = {10.48550/ARXIV.1708.07747},
  publisher = {arXiv},
}

@Article{norcliffe2025stochastic,
  author  = {Norcliffe, Alexander and Lee, Changhee and Imrie, Fergus and Van Der Schaar, Mihaela and Li{\`o}, Pietro},
  journal = {arXiv preprint arXiv:2508.01957},
  title   = {Stochastic Encodings for Active Feature Acquisition},
  year    = {2025},
}

@Book{Miller2002,
  author    = {Miller, Alan},
  publisher = {chapman and hall/CRC},
  title     = {Subset selection in regression},
  year      = {2002},
}

@Article{Kusner2014,
  author    = {Kusner, Matt and Chen, Wenlin and Zhou, Quan and Xu, Zhixiang (Eddie) and Weinberger, Kilian and Chen, Yixin},
  doi       = {10.1609/aaai.v28i1.8967},
  journal   = {Proceedings of the AAAI Conference on Artificial Intelligence},
  number    = {1},
  title     = {Feature-Cost Sensitive Learning with Submodular Trees of Classifiers},
  volume    = {28},
  year      = {2014},
  issn      = {2159-5399},
  month     = jun,
  publisher = {Association for the Advancement of Artificial Intelligence (AAAI)},
}

@InProceedings{jang2016categorical,
  author    = {Eric Jang and Shixiang Gu and Ben Poole},
  booktitle = {International Conference on Learning Representations},
  title     = {Categorical Reparameterization with Gumbel-Softmax},
  year      = {2017},
  url       = {https://openreview.net/forum?id=rkE3y85ee},
}
\bibliographystyle{icml2026}
\clearpage
\onecolumn
\begin{appendices}
  \appendix
\setcounter{equation}{0}
\setcounter{table}{0}
\setcounter{algorithm}{0}
\setcounter{theorem}{0}
\counterwithin*{equation}{section}
\renewcommand\theequation{\thesection.\arabic{equation}}
\renewcommand\thetheorem{\thesection.\arabic{theorem}}

\section{Submodularity of Template Set Optimization}\label{sec:submodularity}
Before showing the submodular property for our proposed template search problem, we first restate the ``facility location assignment'' problem described in \citet{krau-2014}. Further details regarding submodularity and its applications can be found in \citet{krau-2014}.

\paragraph{Facility Location} Suppose we wish to select some locations out of a set $\mathcal{V} \equiv [V]$ to open up facilities in order to serve a collection of $M$ customers. If we open up a facility at location $v$, then it provides a service of value $R_{m,v}$ to customer $m \in \mathcal{M} \equiv [M]$, where $R \in \mathbb{R}^{M \times V}$. If each customer chooses the facility with the highest value, the total value provided to all customers can be modeled with the following set function:
\begin{equation}
    f(\widetilde{\mathcal{V}}) \coloneq \sum_{m \in \mathcal{M}} \max_{\widetilde{v} \in \widetilde{\mathcal{V}}} R_{m,\widetilde{v}}, \qquad f(\varnothing) \coloneq 0
    \label{eq:facility_loc}
\end{equation}
where $\widetilde{\mathcal{V}} \subseteq \mathcal{V}$ is a potential set of locations to open up facilities. If $R_{i,j} \geq 0 \forall i,j$, then $f(\cdot)$ is a monotone submodular function \citep{Frieze1974}.

\begin{theorem}
    (informal) The template collection objective function $g(\mathcal{B})$ defined in \autoref{eq:tafa_set_obj_fn} is submodular.
    \begin{proof}
        To prove our template set optimization has a submodular property, we first rewrite our minimization problem as a maximization one, which is trivial as maximization and minimization problems differ by a negative sign. That is, we define a maximization objective function $h: \mathfrak{S} \rightarrow \mathbb{R}$ such that $\min_{\mathcal{B} \in \mathfrak{S}}g(\mathcal{B}) \Leftrightarrow \max_{\mathcal{B} \in \mathfrak{S}} h(\mathcal{B})$. As we do not have knowledge about the underlying data generation process, we replace the expectation with its empirical counterpart and ignore the constant normalization factor term in computing the average.
        \begin{gather*}
            h(\mathcal{B}) = \sum_{x^{(n)},y^{(n)} \in \mathcal{D}} \left[  
                \max_{\textbf{b} \in \mathcal{B}}  
                -l\left( \widehat{y} \left( x^{(n)}_{\textbf{b}} \right), y^{(n)} \right) - \lambda \sum_{b \in \textbf{b}} c(b)  
                \right] \\  
            h(\varnothing) = 0
        \end{gather*}
        where symbols $g$ and $\mathfrak{S}$ are defined in \autoref{eq:tafa_set_obj_fn} and \autoref{eq:tafa_opt_prob}, respectively. The strategy for proving $h$ has submodular property is, through algebraic manipulation, we show the rewritten version of $h$ has the same form as \autoref{eq:facility_loc}.
        \begin{equation*}
            \begin{aligned}
                h(\mathcal{B}) & =  \sum_{x^{(n)},y^{(n)} \in \mathcal{D}} {                                 
                    \max_{\textbf{b}^{(b)} \in \mathcal{B}}  
                    -l\left( \widehat{y} \left( x^{(n)}_{\textbf{b}^{(b)}} \right), y^{(n)} \right) - \lambda \sum_{b \in \textbf{b}^{(b)}} c(b)  
                }                                                                                            \\ 
                               & \displaystyle_{=}^{{\text{eq. 1}}} \sum_{x^{(n)},y^{(n)} \in \mathcal{D}} {
                    \max_{\textbf{b}^{(b)} \in \mathcal{B}} -e\left( \widehat{y} \left( x^{(n)}_{\textbf{b}^{(b)}} \right), y^{(n)} \right)
                }                                                                                            \\
                               & = \sum_{n \in \mathcal{D}} {
                    \max_{b \in \mathcal{B}} -H_{n,b}
                }
            \end{aligned}\addtocounter{equation}{1}\tag{\theequation} \label{eq:tafa_set_obj_submodular_form}
        \end{equation*}
        where $H \in \mathbb{R}^{\abs{\mathcal{D}} \times \mathcal{B}}$ is a matrix and $H_{n,b} = -e\left( \widehat{y} \left( x^{(n)}_{\textbf{b}^{(b)}} \right), y^{(n)} \right)$ denotes the $n^{\text{th}}$ row and $b^{\text{th}}$ column of $H$.\footnote{We abuse the notation in the following proof and assume that a set is ordered and the $n^{\text{th}}$ element of a set $\mathcal{D}$ is denoted as $n \in \mathcal{D}$.} Observing \autoref{eq:tafa_set_obj_submodular_form} has the same form as \autoref{eq:facility_loc} completes the proof.
    \end{proof}
\end{theorem}

\section{Theoretical Guarantee of Template Greedy Search} \label{sec:greedy_optimality}
In this section, we show that a special choice of task objective and cost function makes \autoref{eq:tafa_set_obj_submodular_form} a monotone submodular set function. \citet{nemh-1978} proves a celebrated result that the greedy algorithm provides a good approximation to the optimal solution to NP-hard optimization problem, which we restate below
\begin{theorem} \label{thm:greedy_thm}
    (\citet{nemh-1978}) Fix a non-negative monotone submodular function $f: 2^{\mathcal{V}} \rightarrow \mathbb{R}_{+}$ and let $\left\{\mathcal{S}_i\right\}_{i \geq 0}$ be the greedily selected sets, where $\mathcal{S}_i = \mathcal{S}_{i-1} \cup \left\{\argmax_{e} \delta\left( e \vert \mathcal{S}_{i-1} \right)\right\}$ and $\Delta$ is the discrete derivative. Then for all positive integers $k$ and $l$,
    \begin{equation*}
        f(\mathcal{S}_t) \geq \left( 1 - e^{-l/k} \right) \max_{\mathcal{S}: \abs{\mathcal{S}} \leq k } f(\mathcal{S})
    \end{equation*}
    In particular, for $l = k$, $f(\mathcal{S}_k) \geq (1 - 1/e) \max_{\abs{S} \leq k} f(\mathcal{S})$.
\end{theorem}
As previously noted in \autoref{sec:submodularity}, for the facility location outline in \autoref{eq:facility_loc}, showing $R_{i,j} \geq 0, \forall i,j$ implies that $f(\cdot)$ is a monotone submodular function \citep{Frieze1974}. Here, we make note of a special case of \autoref{eq:tafa_set_obj_submodular_form} is monotonic and therefore enjoys the property in \autoref{thm:greedy_thm}. Suppose we are designing a policy to solve a classification task; if we choose a $\lambda$-shifted negative zero-one loss as our task loss function, i.e. $l\left(\widehat{y}\left(x_{\textbf{b}}\right), y\right) = -\mathds{1}\left(\widehat{y}\left(x_{\textbf{b}}\right) = y\right) - \lambda$ and a constant function of $\frac{1}{D}$ for all features as the acquisition cost $c(\cdot)$, then $-H_{n,b} = \mathds{1}\left(\widehat{y}\left(x_{\textbf{b}^{(b)}}^{(n)}\right) = y\right) + \lambda - \lambda \sum_{b \in \textbf{b}^{(b)}} \frac{1}{D} \geq 0$ for any $\lambda \in \mathbb{R}_{+}$, which makes this special case a monotone submodular function that enjoys the theoretical approximation guarantee specified in \autoref{thm:greedy_thm}. We acknowledge that many practical choices of task loss and acquisition cost function would not have the monotonicity property; yet, our experiments show that the greedy method could still provide decent performance.

\begin{table*}[t]
    \centering
    \caption{TAFA Method Experimental Configurations}
    \label{tab:tafa-configs}
    \small
    \begin{tabular}{@{}llllll@{}}
        \toprule
        \textbf{Dataset} & \textbf{Architecture} & \textbf{Hidden Units} & \textbf{Activation} & \makecell{\textbf{$\lambda$ range}       \\(low, high, step)} & \textbf{$k$-neighs.} \\
        \midrule
        Big5             & 2-layer FCN           & {[}256, 256{]}        & ReLU                & 0.0, 0.31, 0.02                    & 10  \\
        Cube             & Linear Layer          & ---                   & ---                 & 0.0, 0.31, 0.02                    & 10  \\
        Gas              & 2-layer FCN           & {[}256, 256{]}        & ReLU                & 0.0, 0.31, 0.02                    & 10  \\
        Grid             & 2-layer FCN           & {[}128, 128{]}        & ReLU                & 0.0, 1.161, 0.02                   & 10  \\
        MNIST            & 2-layer FCN           & {[}512, 512{]}        & ReLU                & 0.05, 0.0751, 0.0005               & 100 \\
        Fashion          & 2-layer FCN           & {[}512, 512{]}        & ReLU                & 0.05, 0.0751, 0.0005               & 100 \\
        \bottomrule
    \end{tabular}
\end{table*}
\section{Upper Bounds on TAFA Policy Value Function}\label{sec:tafa_value_bound}
It is worth mentioning that \autoref{eq:feature_acquisition} is upper bounded by its ACO counterpart and is further upper bounded by its ACO counterpart. Let $V^{t}$ denote the state value function of an AFA MDP at the $t^{\text{th}} \in 1 \dots T$ step, where $T$ is the maximum number of acquisitions an agent is allowed to make, i.e.
\begin{equation*}
    \begin{gathered}
        V^{0}(x_{\textbf{o}}) \coloneq -\mathbb{E}_{y \vert x_{\textbf{o}}}{\left[
                    l\left(\widehat{y}\left(x_{\textbf{o}}\right), y\right)
                    \right]} \\
        V^{t}(x_{\textbf{o}}) \coloneq \max{ \left( V^{0}\left( x_{\textbf{o}} \right), \max_{d \in [D] \backslash \textbf{o}}{\left( -\lambda c(d) + \mathbb{E}_{x_d \vert x_{\textbf{o} \cup \left\{ d \right\}}}\left[ l\left(\widehat{y}\left(x_{\textbf{o} \cup \left\{ d \right\}}\right), y\right) \right] \right)} \right)} \\
        \text{ACO}^{t}(x_{\textbf{o}}) \coloneq \max_{\textbf{u} \subset \mathcal{A}}{-\mathbb{E}_{y, x_{\textbf{u}} \vert x_{\textbf{o}}}\left[ l\left(\widehat{y}\left(x_{\textbf{o} \cup \textbf{u}}\right), y\right) \right] - \lambda \sum_{u \in \textbf{u}} c(u)}
    \end{gathered}
\end{equation*}
where $\mathcal{A} = \left\{ \textbf{u} \vert \textbf{u} \subset [D] \backslash \textbf{o} \text{ and } \abs{\textbf{u}} \leq T\right\}$ is the collection of all possible available action sequences for the rest of the ACO rollout. By Thm. C.1 of \citet{vala-2024} and because our TAFA template collection $\mathcal{B} \subseteq \mathcal{A}$, it is trivial to show that
\small
\begin{align*}
    V^{t}(x_{\textbf{o}}) & \geq \text{ACO}^{t}(x_{\textbf{o}})                                                                                                                                                                                       \\
                          & \geq \max_{\textbf{b} \in \mathcal{B}}{-\mathbb{E}_{y, x_{\textbf{b}} \vert x_{\textbf{o}}}\left[ l\left(\widehat{y}\left(x_{\textbf{o} \cup \textbf{b}}\right), y\right) \right] - \lambda \sum_{u \in \textbf{b}} c(u)}
\end{align*}
\normalsize

\section{Experimental Setup} \label{sec:implementation}
All experiments were conducted on a Lenovo ThinkStation P520 workstation equipped with an Intel Xeon W-2295 18-core processor running at 3.0 GHz (4.8 GHz boost) and 512 GB of DDR4 ECC memory. The system runs Ubuntu 24.04 LTS with Python 3.13.2.

All algorithms were implemented using single-threaded execution to ensure fair comparison across methods. The datasets used in our experiments are publicly available and were preprocessed according to standard practices in the literature. Both the neural network architecture and the trade-off parameters used in our experiments are shown in \autoref{tab:tafa-configs}. For all baselines, we use the official implementation in their respective GitHub repositories.

All timing measurements exclude data loading and preprocessing time, focusing solely on the algorithm execution time. Memory usage was monitored throughout the experiments to ensure no memory constraints affected the results.

For the results of \texttt{tafa-interp} in \autoref{fig:acc_vs_nfeats}, we train an ensemble of trees with each tree's leaf limit of 4 (big5), 4 (cube), 16 (gas), 8 (grid), 8 (mnist), 8 (fashion). For the DDT and SYMPOL baselines, we use the official implementations provided in the respective papers' GitHub repositories. To adapt DDT and SYMPOL for AFA tasks, we define the goal of the RL as maximizing the predictive loss while minimizing the acquisition cost: $\mathrm{reward}=-l\left(\widehat{y}\left(x_{\textbf{o}}\right), y\right) - \lambda \sum_{u \in \textbf{o}} c(u)$, following \citet{li-2021}. For consistent results, DDT, SYMPOL, and \texttt{tafa-interp} are trained with the same classifier $\widehat{y}$ and fixed $\lambda$. To quantify the tree’s policy interpretability, we follow the common practice in RL interpretable literature by using reward as the metric and leaf count as a proxy for rule complexity, as fewer leaves would imply shorter paths and simpler rules \citep{roth2019conservative, glanois2024survey}.

\section{Additional Interpretability Result} \label{sec:interp-appx}
\subsection*{Automotive Diagnosis}\label{sec:engine-interp}
In the this section, to showcase TAFA's applicability in interpretability-critical domains, we demonstrate how one can apply TAFA framework to make intelligent choices of feature subsets to diagnose car engines recorded in the EngineDB dataset~\citep{verg-2023}. We use decision trees as predictors and employ only iterative mutate template search \autoref{alg:mutate_greedy_search} forgoing continuous refinement \autoref{alg:training_procedure}. This configuration is appropriate for three reasons: First, automotive diagnostics often requires fully transparent decision logic for regulatory compliance and technician understanding. Decision trees provide this interpretability but are non-differentiable, precluding gradient-based template optimization. Second, the EngineDB dataset~\citep{verg-2023} (D=11 features, structured sensor measurements) has lower dimensionality and stronger domain structure than the high-dimensional problems (e.g., Fashion-MNIST, D=256) where continuous refinement provides substantial gains (5-8\% improvement, \autoref{fig:mutation-progres}~(top)). Our ablation studies suggest genetic search alone suffices for such structured, moderate-dimensional spaces. Third, the quality of the discovered templates validates this choice: as analyzed below, the templates acquire diagnostically meaningful feature combinations (e.g., {RPM, Speed, CO, O2} for low-speed conditions) that align with automotive engineering principles, without requiring gradient optimization. This demonstrates TAFA's modularity: practitioners can select predictor and optimization strategy based on domain constraints (interpretability, compute resources, dimensionality) without abandoning the template-based framework. The genetic search remains effective when the problem structure is well-defined, as we show below.

Here, the task for our TAFA agent is to sequentially gather information to diagnose a car engine that could be in one of four states: \texttt{normal} engine, running \texttt{Lean}/\texttt{rich}, or having an ignition \texttt{voltage} issue. See \citet{verg-2023} for further details. We choose average fuel \texttt{consumption} (L/100km) as the initial feature $o_{\text{init}}$ since car owners often track their gas mileage and would notice it when ones start spending more on gasoline. We provide a brief background knowledge about failure mode recorded in EngineDB dataset~\citep{verg-2023}; further details can be found at \citet{verg-2023}.
\begin{itemize}
    \item An engine running \texttt{rich} means it receives too much fuel relative to the amount of air in the combustion chamber, resulting in incomplete combustion that produces high carbon monoxide (CO) emissions, reduced fuel efficiency, and potential engine damage from fuel dilution of engine oil.
    \item An engine running \texttt{lean} occurs when there's too little fuel relative to air, causing higher combustion temperatures that can lead to engine knocking, overheating, and potential damage to pistons and valves, while also producing distinctive oxygen signatures in exhaust gases
    \item A \texttt{voltage} issue in the ignition circuit means the spark plugs receive insufficient electrical power to properly ignite the fuel-air mixture, resulting in misfires, rough idling, reduced power output, and poor fuel economy due to incomplete or delayed combustion
\end{itemize}

\begin{figure*}[tbp]
    \centering
    \begin{subfigure}{0.9\linewidth}
        \begin{mdframed}[linewidth=1pt, roundcorner=3pt, backgroundcolor=gray!5, linecolor=black]
            \begin{tcolorbox}[colback=blue!5!white, colframe=blue!50!black, title=\textbf{Acquire Set 1 (Freq: 53.5\%)}]
                \textbf{Features:} $\{\texttt{Consumption}, \texttt{CO}\}$ \\
                \textbf{Rule 1.1:} If $7.1335 < \texttt{Consumption} \leq 13.1900$ (L/100km) \\
                \vspace{0.1cm}
                \small
                \begin{tabular}{@{}cccc@{}}
                    \texttt{Normal} & \texttt{Rich} & \texttt{Lean} & \texttt{Voltage} \\
                    38.3\%          & 20.4\%        & 20.1\%        & 21.2\%
                \end{tabular}
                \normalsize
            \end{tcolorbox}
            \begin{tcolorbox}[colback=blue!5!white, colframe=blue!50!black, title=\textbf{Acquire Set 2 (Freq: 28.0\%)}]
                \textbf{Features:} $\{\texttt{Consumption}, \texttt{Speed}, \texttt{O2}\}$ \\
                \textbf{Rule 2.1:} If $\texttt{Consumption} \leq 6.6905$ (L/100km) \\
                \vspace{0.1cm}
                \small
                \begin{tabular}{@{}cccc@{}}
                    \texttt{Normal} & \texttt{Rich} & \texttt{Lean} & \texttt{Voltage} \\
                    19.1\%          & 12.9\%        & 36.6\%        & 31.4\%
                \end{tabular}
                \normalsize
            \end{tcolorbox}
            \begin{tcolorbox}[colback=blue!5!white, colframe=blue!50!black, title=\textbf{Acquire Set 3 (Freq: 10.4\%)}]
                \textbf{Features:} $\{\texttt{Consumption}, \texttt{HC}\}$ \\
                \textbf{Rule 3.1:} If $\texttt{Consumption} > 13.1900$ (L/100km) \\
                \vspace{0.1cm}
                \small
                \begin{tabular}{@{}cccc@{}}
                    \texttt{Normal} & \texttt{Rich} & \texttt{Lean} & \texttt{Voltage} \\
                    0.0\%           & 34.4\%        & 32.3\%        & 33.3\%
                \end{tabular}
                \normalsize
            \end{tcolorbox}
            \begin{tcolorbox}[colback=blue!5!white, colframe=blue!50!black, title=\textbf{Acquire Set 4 (Freq: 3.3\%)}]
                \textbf{Features:} $\{\texttt{RPM}, \texttt{Consumption}, \texttt{Speed}, \texttt{CO}, \texttt{O2}\}$ \\
                \textbf{Rule 5.1:} If $6.6905 < \texttt{Consumption} \leq 7.1335$ (L/100km) and $\texttt{Speed} \leq 41.6110$ (kph) \\
                \vspace{0.1cm}
                \small
                \begin{tabular}{@{}cccc@{}}
                    \texttt{Normal} & \texttt{Rich} & \texttt{Lean} & \texttt{Voltage} \\
                    48.3\%          & 44.4\%        & 3.4\%         & 3.9\%
                \end{tabular}
                \normalsize
            \end{tcolorbox}
            \begin{tcolorbox}[colback=blue!5!white, colframe=blue!50!black, title=\textbf{Acquire Set 5 (Freq: 4.7\%)}]
                \textbf{Features:} $\{\texttt{Force}, \texttt{RPM}, \texttt{Rate}, \texttt{Consumption}, \texttt{Speed}, \texttt{CO}, \texttt{O2}\}$ \\
                \textbf{Rule 4.1:} If $6.6905 < \texttt{Consumption} \leq 7.1335$ (L/100km) and $\texttt{Speed} > 41.6110$ (kph) \\
                \vspace{0.1cm}
                \small
                \begin{tabular}{@{}cccc@{}}
                    \texttt{Normal} & \texttt{Rich} & \texttt{Lean} & \texttt{Voltage} \\
                    15.2\%          & 2.7\%         & 43.0\%        & 39.1\%
                \end{tabular}
                \normalsize
            \end{tcolorbox}
        \end{mdframed}
        \caption{Texted based representation of the extracted rules.}
        \label{fig:posthoc_engine_text}
    \end{subfigure}
    \hfill
    \begin{subfigure}{.6\linewidth}
        \includegraphics[width=\linewidth]{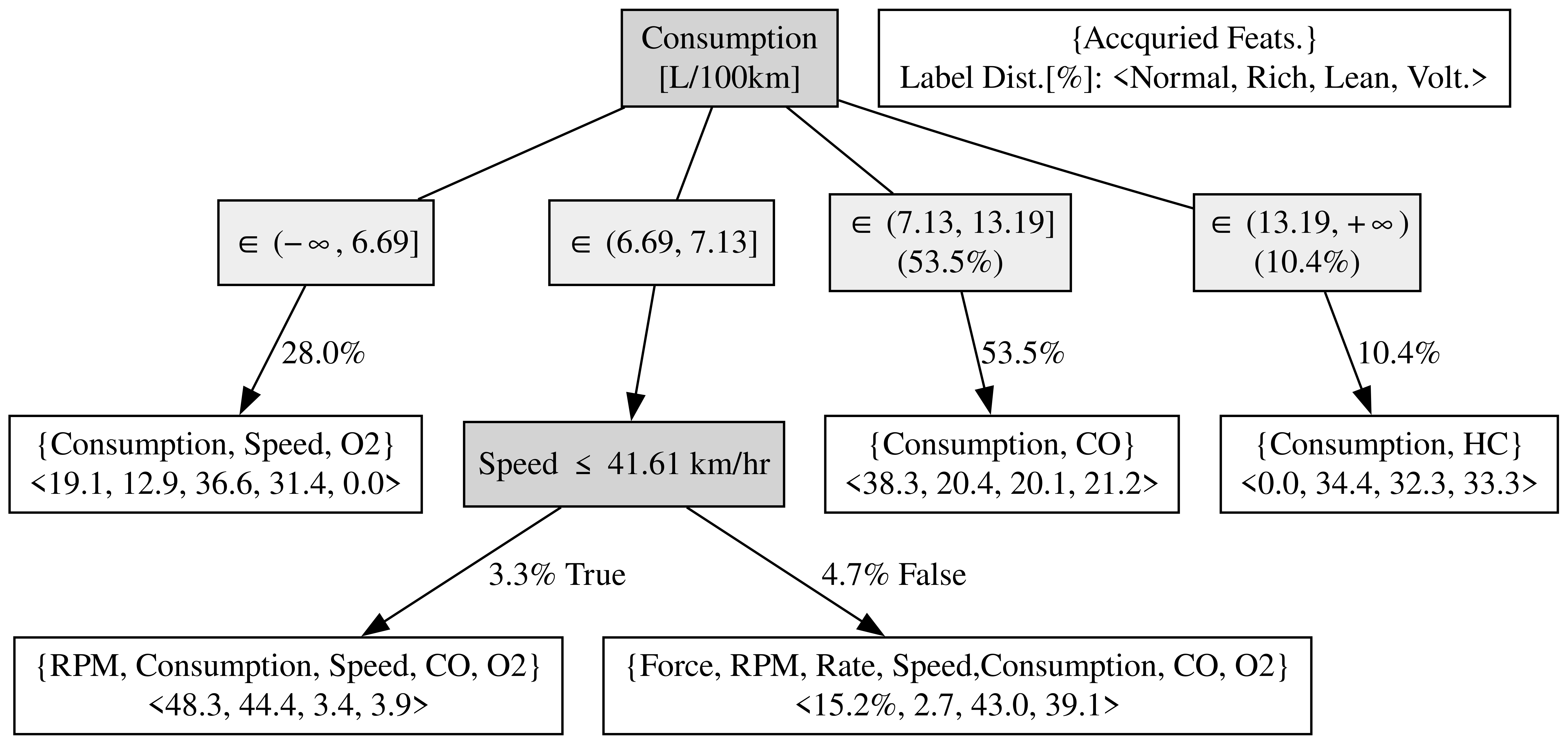}
        \caption{Tree based representation of the extracted rules, meaning this figure represents the same information as that of in \autoref{fig:posthoc_engine_text}.}
        \label{fig:posthoc_engine}
    \end{subfigure}
    \hfill
    \caption{Engine fault diagnosis rules extracted by TAFA in both text format and in tree format.}
    \label{fig:engine-rules}
\end{figure*}

Below, we analyze two acquisition sets---set 4 and set 5---from \autoref{fig:engine-rules} that handle mild to slightly elevated fuel consumption (6.69-7.13 L/100km), which can be challenging to diagnose. Additionally, \autoref{fig:engine-rules} shows the interpretable rules extracted from our TAFA policy, whereas \autoref{fig:engine-subset-dtc-set4} and \autoref{fig:engine-subset-dtc-set5} show the downstream interpretable predictor that acts on the features selected with our TAFA policy.

\paragraph{Acquire Set 4 (Low Speed, $< 41.61$ kph)} Cases assigned to use this set of rule are being operated in low-speed regime and are experiencing average or slightly low fuel efficiency; for thsee cases, our policy decides to acquire \texttt{RPM}, \texttt{Speed}, \texttt{CO}, and \texttt{O2} in addition to the initial feature (fuel \texttt{consumption}). From \autoref{fig:engine-rules}, we see that majority of cases that get assigned to use this set of features are either \texttt{normal} or \texttt{rich}, and there are with very little cases being \texttt{lean} or \texttt{voltage}. As previously observed in \citet{verg-2023}, rich-running engines often have very distinctive pattern in terms of chemicals released from the emission, which naturally explains our policy's decision to acquire both \texttt{CO} and \texttt{O2} at the same time. Besides the obvious choice, we think our policy's decision to acquire \texttt{RPM} especially interesting. Recall cases assigned to this rule are all in low speed regime, i.e., below 41.61 kph, suggesting the vehicle is highly likely being stuck in city stop-and-go traffic where big throttle input is often applied to get the vehicle up to speed as the traffic light turns green. In this scenario, the engine revs up as the vehicle accelerate, meaning more gasoline is being burned and the tacometer reading (measured in \texttt{RPM}) is going to increase. In this case, a car having above average gas mileage could be due to the condition it was operated in, not because of any malfunction. From the visualization (\autoref{fig:engine-subset-dtc-set4}) of our predictor using the chosen set of features, one can see that, among the three leaves that classifies an engine being \texttt{normal}, the ascendants of two of the three leaves uses \texttt{RPM} as one of the feature, suggesting the importance of \texttt{RPM} at distinguishing \texttt{normal} running engine, and further verifies the intelligence of our TAFA policy.
\begin{figure}
    \centering
    \includegraphics[width=\linewidth]{interp-engine/set4.png}
    \caption{The decision tree calssifier using set 4.}
    \label{fig:engine-subset-dtc-set4}
\end{figure}

\paragraph{Acquire Set 5 (High Speed, $> 41.61$ kph)} In contrast to set 4, all cases assigned to this set are all in the high speed regime with average or slightly low fuel efficiency; regarding the features acquired, in addition to all the features in Set 4, this set additionally acquire rotational \texttt{force} (measuring the fuel-to-power conversion efficiency) and short-term fuel \texttt{rate} (in L/hr). Our policy's decision to acquire these two addtional features, again, makes sense: at highway speeds, healthy engines generate strong force while faulty engines produce weak power despite having similar mild or slightly above average fuel consumption. Here, inspecting the predictor (\autoref{fig:engine-subset-dtc-set5}), we again see both \texttt{force} and \texttt{rate} play an essential role at correctly diagnosing car engines, especially \texttt{force}; in the middle branch of \autoref{fig:engine-subset-dtc-set5}, \texttt{force} is the second feature that the decision tree classifier splits on, showing the imporance of \texttt{force} in diagnosing vehicles in high speed regime expeincing mild or slightly above average fuel efficiency.
\begin{figure}
    \centering
    \includegraphics[width=\linewidth]{interp-engine/set5.png}
    \caption{The decision tree acquired using set 5.}
    \label{fig:engine-subset-dtc-set5}
\end{figure}

Both acquisition strategies demonstrate TAFA's ability to learn interpretable diagnostic strategies aligned with automotive engineering principles. We again stress that all the pattern observed above are all completely learned from data, and none of the prior knowledge described above (with the exception of choosing \texttt{consumption} as initial feature) were known to the TAFA policy at any point during training. Despite using a non-differentiable decision tree classifier as a predictor and therefore not using gradient based template fine-tuning, our study shows TAFA policy is still able to identify high-quality feature templates and allow users of our framework to gain useful insights into the reasoning made by our TAFA policy. This demonstrates TAFA's modularity: practitioners can select predictor and optimization strategy based on domain constraints (interpretability, compute resources, dimensionality) without abandoning the template-based framework. The genetic search outlined in \autoref{alg:mutate_greedy_search} remains effective as we just shown above with the automotive diagnosis example.

\end{appendices}

\end{document}